\documentclass[11pt]{article}

\usepackage[final]{acl}

\usepackage{times}
\usepackage{latexsym}
\usepackage{paralist}
\usepackage{inconsolata}
\usepackage{algorithm}
\usepackage{algorithmic}
\usepackage{amsmath}
\usepackage{booktabs}
\usepackage{multirow}
\usepackage{array}
\usepackage{enumitem}
\usepackage{listings}

\usepackage[T1]{fontenc}

\usepackage[utf8]{inputenc}

\usepackage{microtype}

\usepackage{tikz}
\usetikzlibrary{arrows.meta,positioning,decorations.pathreplacing}

\usepackage{graphicx}

\title{INSURE-Dial: A Phase-Aware Conversational Dataset \& Benchmark for Compliance Verification and Phase Detection}

\author{
  Shubham Kulkarni$^{1}$ \quad
  Alexander Lyzhov$^{2}$ \quad
  Preetam Joshi$^{2}$ \quad
  Shiva Chaitanya$^{1}$\\
  $^{1}$Interactly.ai, California, USA \\
  $^{2}$AIMon Labs, California, USA \\
  \texttt{shubham@interactly.ai, alex@aimon.ai,  preetam@aimon.ai, shiva@interactly.ai}
}

\usepackage{fvextra}

\DefineVerbatimEnvironment{PromptBlock}{Verbatim}{
  fontsize=\scriptsize,
  breaklines=true,
  breakanywhere=true,
  breaksymbol={},        
  breaksymbolleft={},
  breaksymbolright={},
  breaksymbolsep=0pt,
  breakindent=0pt,       
  breakautoindent=false, 
  xleftmargin=0pt        
}
\begin{document}
\maketitle
\begin{abstract}
Administrative phone tasks drain roughly \$1 trillion annually from U.S. healthcare, with over 500 million insurance–benefit verification calls manually handled in 2024. We introduce \textbf{INSURE-Dial}, to our knowledge the first public benchmark for developing and assessing \emph{compliance-aware} voice agents for phase-aware call auditing with span-based compliance verification. The corpus includes 50 de-identified, AI-initiated calls with live insurance representatives (mean 71 turns/call) and 1{,}000 synthetically generated calls that mirror the same workflow. All calls are annotated with a phase-structured JSON schema covering IVR navigation, patient identification, coverage status, medication checks (up to two drugs), and agent identification (CRN), and each phase is labeled for \emph{Information} and \emph{Procedural} compliance under explicit ask/answer logic. We define two novel evaluation tasks: (1) \emph{Phase Boundary Detection} (span segmentation under phase-specific acceptance rules) and (2) \emph{Compliance Verification} (IC/PC decisions given fixed spans). Per-phase scores are strong across small, low-latency baselines, but end-to-end reliability is constrained by span-boundary errors. On real calls, full-call exact segmentation is low, showing a gap between conversational fluency and audit-grade evidence.

\end{abstract}
\section{Introduction}
Administrative phone calls remain a major hidden cost in healthcare, accounting for approximately 25\% of the \$4.5 trillion annual expenditure in the U.S., nearly \$1 trillion each year~\cite{mckinsey2024weathering, lavoie2025artificial}. Insurance benefit verification alone accounted for more than 500 million calls in 2024, significantly contributing to a total of 3.2 billion annual healthcare-related calls, more than 90\% of which remain fully manual~\cite{CAQH2024Index}.
These calls typically require healthcare personnel to navigate complex automated phone menus (interactive voice response; \textbf{IVR}), endure lengthy waiting times (frequently exceeding 50 minutes), and repeatedly provide or request the same critical patient data~\cite{Landi2025VoiceCareAI}. In practice, organizations audit these conversations against \emph{checklists} that capture required informational and procedural steps (e.g., verifying patient identity before discussing coverage, documenting coverage status, and recording the insurer representative’s identifier). Such steps are motivated by privacy and compliance regimes (e.g., HIPAA)~\cite{ecfr164514,proofpoint2024ai} and broader AI-governance expectations for traceability and transparency (e.g., the EU AI Act)~\cite{walters2023complyingeuaiact}.
Recent advances in large language models (LLMs), automatic speech recognition, and neural synthesis now make it feasible to automate complex, multi-step healthcare dialogues at scale~\cite{adams2025generative,Stapleton2025SuperDialBlueprint}. In practice, providers are beginning to deploy \emph{AI agents} that place outbound calls on behalf of clinical staff to verify pharmacy benefits with insurers; navigating IVRs, confirming coverage status, checking up medications per patient, and logging the representative’s identity under an ordered audit checklist and HIPAA-related rules.
This checklist view follows our prior phase-level auditing formalism, \emph{Obligatory-Information Phase Structured Compliance Evaluation (OIP--SCE)}~\cite{kulkarni2026all}, which represents each auditable obligation as a phase with an explicit start/finish, a verdict, and a minimal evidence trail, and evaluates calls under conjunctive \emph{coverage} and \emph{order-safety} requirements. 
We use the term \emph{phase} to mean a \textbf{contiguous span of turns} corresponding to a single audit-relevant checklist item (e.g., \emph{Patient Identification} or \emph{Coverage Status}). This definition is consistent with the \emph{obligatory-information phase} abstraction introduced in OIP--SCE, which treats regulated dialogue compliance as conjunctive, order-sensitive progress through a phase inventory with auditable evidence. Phases provide an intermediate structure between turn-level labels and whole-call outcomes, enabling evaluation of \emph{what happened, where it happened}, and \emph{who asked/answered} within the transcript. Consider a benefit-verification call in which an auditor asks: \emph{“Did the agent verify the member’s identity before discussing coverage?”} This requires locating the \emph{Patient Identification} phase (a turn span such as turns 6--10 where the agent requests DOB and the representative confirms it) and then verifying whether required fields were obtained and whether the correct party provided them. A whole-call ``success'' label or a generic dialogue quality score cannot provide this localized, checklist-aligned evidence.
Audits require structured evidence that each required step occurred (often with ordering constraints), with the correct asker/answerer and unambiguous content; subtle omissions can create compliance risk~\cite{hetrick2023chatgpt}. We do \emph{not} claim to solve automated auditing end-to-end; rather, we introduce a benchmark that measures the gap between current model behavior and audit-style requirements.
Existing NLP benchmarks largely target relevance, helpfulness, or generic safety and governance criteria~\cite{pool2024large,fan2024goldcoin,freyer2024future,pasch2025balancing,ferreira2025llm,iso27001_2022,soc2_aicpa,iso23894_2023,owasp2025agentic,liu2024shield}. They do not test whether a transcript contains \emph{checklist-aligned phases} with recoverable spans under workflow rules. Free-form LLM summaries are also insufficient for auditing because they are not reliably deterministic or verifiable at the span level~\cite{abbasian2024foundation}. Building on OIP--SCE’s phase-level, order-aware auditing formalism, we address this gap by introducing INSURE-Dial, to our knowledge the first public benchmark for \emph{insurance benefit-verification calls} that operationalizes phase-aware compliance evaluation as two explicit tasks: (i) boundary recovery and (ii) rule-based compliance given fixed spans.
INSURE-Dial contains (i) 50 de-identified real calls and (ii) 1{,}000 synthetic calls generated with \textbf{Dataframer} to expand coverage of common flows and corner cases (e.g., inactive plans; one- vs.\ two-drug workflows). Each call is annotated with a phase-aware JSON schema (\ref{sec:annotation}) that records phase spans and distinguishes \emph{information compliance} (was required information obtained) from \emph{procedural compliance} (was the interaction carried out correctly under workflow rules), including explicit ask/answer roles. We define two benchmark tasks: \emph{Phase Boundary Detection} (recover phase spans under phase-specific acceptance rules) and \emph{Compliance Verification} (predict compliance labels given fixed spans). Our experiments suggest that models can judge compliance reliably when spans are correct; however, span-boundary errors substantially reduce end-to-end auditability. We quantify this gap with these two tasks. We release the schema, prompts, scoring code, and de-identified data.\footnote{\url{https://github.com/interactly-aimon/insuredial-dataframer-eacl-2026}; dataset: \url{https://huggingface.co/datasets/dataframer/insure-dial}.}
\section{Use\mbox{-}Case and Annotation Framework}
\label{sec:annotation}
We study a single operational workflow common in U.S. tele-pharmacy: verifying a patient’s pharmacy benefits for up to two medications over the phone. We instrument this workflow for post-call auditing. Given a transcript, an \emph{extractor LLM} produces a structured JSON \emph{audit record} whose top-level keys follow a fixed phase order (Appendix Fig.~\ref{fig:app_framework}). Each phase stores (i) an \emph{inclusive} turn span \(S=[i,j]\) and, when applicable, (ii) role-typed ask/answer flags \((A,B)\) and (iii) minimal content fields (e.g., coverage status, restrictions, copay/coinsurance). We refer to this JSON as the \emph{reference} annotation when scoring: it is human-validated on the real subset (\ref{sec:qa}) and program-validated on the synthetic subset (\ref{sec:synthetic-data-generation}). Model outputs at evaluation time are \emph{predictions}. This phase order reflects a standard operational benefit-verification checklist used in practice for pharmacy benefits calls; while not mandated by a single regulation, it captures the ordered, auditable obligations that human QA teams routinely verify.

\paragraph{Phase inventory:}
A \emph{phase} is a contiguous, role-typed span corresponding to one audit checklist item. We use:
\begin{enumerate}\itemsep0pt
  \item \textbf{Interactive Voice Response (IVR)}: automated menu before a human answers (span only).
  \item \textbf{Greeting (GRT)}: initial assistant--representative salutation (span only).
  \item \textbf{Patient Identification (PID)}: representative requests identifiers; assistant provides them; insurer locates the record.
  \item \textbf{Coverage Status Verification (CSV)}: assistant asks whether the plan is active; representative answers \texttt{ACTIVE}/\texttt{INACTIVE}.
  \item \textbf{Drug loop} (up to two drugs): \textbf{Drug Formulary Verification (DFV)}, \textbf{Drug Restrictions Check (DRC)}, \textbf{Drug Copay/Coinsurance Check (DCC)}.
  \item \textbf{Agent Interaction (CRN)}: assistant requests and records the representative’s name (and optionally a call reference number) for audit logging.
\end{enumerate}
\paragraph{Dialogue flow and gating:}
The intended flow is:
\noindent\(\text{IVR}\!\to\!\text{GRT}\!\to\!\text{PID}\!\to\!\text{CSV}\!\to\![\text{DFV}(k)\!\to\!\text{DRC}(k)\!\to\!\text{DCC}(k)]_{k\in\{0,1\}}\!\to\!\text{CRN}\)\\
We gate downstream phases deterministically:
\begin{enumerate}\itemsep0pt
  \item If \verb|patient_info.record_found=false|, then \textbf{CSV} and the drug loop are \texttt{NA}.
  \item If \verb|coverage_check.plan_status| \verb|!= "ACTIVE"|, then the entire drug loop is \texttt{NA}.
  \item Within a drug: if \verb|formulary.value != YES| then \textbf{DRC} and \textbf{DCC} are \texttt{NA}; if \verb|restrictions.value != NO| then \textbf{DCC} is \texttt{NA}.
\end{enumerate}
In the released audit JSON, drug slot 1 is \texttt{drugs[0]} and drug slot 2 is \texttt{drugs[1]}.
\paragraph{Notation and labels:}
\(S=[i,j]\) denotes an inclusive span with \(i\le j\); \texttt{null} denotes absence; \texttt{NA} denotes inapplicability due to gating. \(A\) and \(B\) are ask/answer flags; roles differ only in \textbf{PID} (representative asks; assistant answers). We score two labels per phase: \emph{Information Compliance (IC)} requires that the required information was explicitly stated; \emph{Procedural Compliance (PC)} additionally requires an ask-then-answer pattern. Spans make these checks auditable and deterministic at the phase level without replaying audio.

\subsection{Quality\mbox{-}Assurance Pipeline}
\label{sec:qa}
\paragraph{Quality-assurance pipeline:}
We pair human review with programmatic checks so that phase boundaries and content fields are reliable before evaluation. Two trained annotators independently label every call; disagreements are adjudicated by a senior reviewer. Agreement on the 50 double-annotated real calls is high (Cohen’s \(\kappa=0.95\pm0.02\)). A rule-based checker validates every JSON by enforcing: (i) span well-formedness and non-crossing constraints; (ii) gating logic; (iii) role/flag consistency; and (iv) schema shape and value constraints (Appendix~\ref{app:checker}).

\textbf{Data collection and de-identification:} From Jan--Mar 2025 we collected ~\(20{,}000+\) operational, business-to-business benefit-verification calls made on behalf of a partner health system; standard recording notifications were in place, and organizational consent was obtained to release a de-identified subset for research. We selected 50 calls via stratified sampling over outcome and flow complexity (Appendix), yielding \textbf{INSURE-Dial-Real} (mean \(71.2\pm38.2\) turns/call; 3{,}559 turns; 42{,}969 tokens); remaining calls are retained internally on encrypted, access-controlled systems and are not released. Audio was transcribed with Whisper-large-v3 \cite{radford2022whisper} and spot-checked. Transcripts were de-identified following HIPAA Safe Harbor (45 CFR \S164.514(b)(2)): direct identifiers (names, member IDs, addresses/phones, precise dates) were removed or masked prior to annotation; only state-level geography is retained.
\section{Synthetic Data Generation}
\label{sec:synthetic-data-generation}
\begin{figure}[t]
    \centering
    \includegraphics[width=\linewidth]{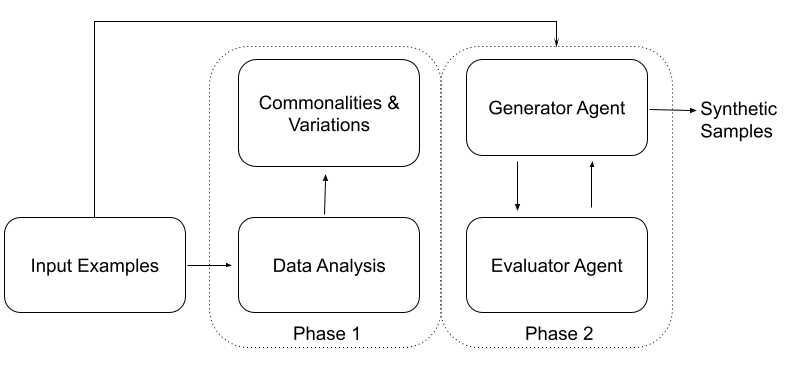}
    \caption{Synthetic data pipeline with an \emph{analysis} stage and a \emph{generation–evaluation} loop.}
    \label{fig:datagen-pipeline}
\end{figure}
Large, well\mbox{-}annotated corpora are essential for robust evaluation but are costly to assemble in regulated domains. Our \textbf{Dataframer}\footnote{\url{https://dataframer.ai}} platform expands a small seed set into a balanced corpus by (i) inferring a lightweight specification of the seed distribution and (ii) generating new samples that meet this specification while maintaining privacy and structure.
\subsection{Pipeline Overview}
Figure~\ref{fig:datagen-pipeline} shows two stages. The \textbf{analysis} stage distills constraints and variation axes from the seed samples along with any additional domain-specific properties provided directly by human experts. The \textbf{generation} stage then alternates a candidate generator and a critic/evaluator in a short feedback loop, accepting a sample once it is judged statistically consistent with the seed set. Decoupling analysis from generation lets a specification generation from a single analysis feed many generation jobs.\newline
\textbf{Phase~I: Analysis:}
Given seed examples, we subsample to fit into the context window and prompt an LLM (e.g., DeepSeek R1~\cite{deepseekai2025deepseekr1incentivizingreasoningcapability} or V3~\cite{deepseekai2024deepseekv3technicalreport}) to return
\((\mathcal{C},\mathcal{V})\), where \(\mathcal{C}=\{c_j\}_j\) is a set of textual constraints (properties that hold across seeds) and
\(\mathcal{V}=\{(a,\mathcal{P}_a)\}\) maps each variation axis / attribute \(a\) to possible values \(\mathcal{P}_a\) it can take in different data samples.
Depending on the analysis settings, \(\mathcal{P}_a\) can extrapolate axes and possible values, or be constrained to those inferred from seeds.\newline
\textbf{Phase~II: Generation:}
Inputs are the seeds, optional \((\mathcal{C},\mathcal{V})\), the target sample count, and parameters. For each requested sample \(s^{(n)}\) we maintain a workspace \(\Omega^{(n)}\) containing few\mbox{-}shot exemplars, target attributes (if sampling over \(\mathcal{V}\)), a history of generator drafts \(\langle g^{(n)}_t\rangle\), and evaluator feedback \(\langle f^{(n)}_t\rangle\).
For up to \(T\) rounds:\newline
\textbf{1. Generation:} Produce \(g^{(n)}_{t}\) to be indistinguishable from seeds and to satisfy target attributes (when present). On later rounds, minimally revise the previous draft per feedback.
\textbf{2. Evaluation:} An LLM emits a binary verdict \(y^{(n)}_{t}\in\{0,1\}\) (pass/fail) and actionable feedback \(f^{(n)}_{t}\) if \(y^{(n)}_{t}=0\), identifying unmet constraints and concrete fixes.
\textbf{3. Loop control:} If \(y^{(n)}_{t}=1\), accept \(g^{(n)}_{t}\) as \(s^{(n)}\); else, append \(f^{(n)}_{t}\) to \(\Omega^{(n)}\) and iterate (optionally including feedback history).\newline
Each accepted synthetic call includes a programmatically validated reference audit JSON (schema of §\ref{sec:annotation}). The JSON must get a pass from a deterministic checker checking schema, spans, gating, roles.
Samples failing validation are discarded.
We report real and synthetic results separately and interpret synthetic primarily for ranking and error patterns, not absolute rates.\newline
\textbf{API and Implementation Details:}
Both stages are available as REST endpoints returning task IDs for orchestration. The API provides: (i) analysis start/status; (ii) generation start/status; (iii) editable prompt templates; and (iv) an interface to review and edit \((\mathcal{C},\mathcal{V})\) before generation, allowing domain experts to refine constraints without modifying code.
\subsection{Dataset Composition}
\label{sec:dataset-stats}
INSURE\mbox{-}Dial contains 50 real and 1{,}000 synthetic calls (total 1{,}050), amounting to 48{,}191 turns (618{,}407 tokens). Table~\ref{tab:stats} summarizes corpus statistics. Synthetic calls mirror real\mbox{-}call structure (phase order, gating, and two\mbox{-}drug loop) while broadening coverage of less common outcomes to support stress\mbox{-}testing downstream extractors.

\begin{table}[t]
\centering
\footnotesize
\resizebox{\columnwidth}{!}{%
\setlength{\tabcolsep}{4pt}
\begin{tabular}{lccc}
\toprule
\textbf{Metric} & \textbf{Real} & \textbf{Synthetic} & \textbf{Total} \\
\midrule
\# Calls & 50 & 1,000 & 1,050 \\
\# Turns & 3,559 & 44,632 & 48,191 \\
Avg.\ Turns/Call & 71.2 & 44.6 & 45.9 \\
Turn Range & 11--215 & 26--70 & 11--215 \\
\# Tokens & 42,969 & 575,438 & 618,407 \\
Drug Queries & 100 & 2,000 & 2,100 \\
Formulary Checks & 27 & 654 & 681 \\
Drugs Covered & 4 (14.8\%) & 466 (71.3\%) & 470 (69.0\%) \\
Records Found & 23 (46.0\%) & 486 (48.6\%) & 509 (48.5\%) \\
Coverage Confirmed & 14 (28.0\%) & 308 (30.8\%) & 322 (30.7\%) \\
\bottomrule
\end{tabular}%
}
\caption{INSURE\mbox{-}Dial corpus statistics across real and synthetic subsets.}
\label{tab:stats}
\end{table}

The real subset exhibits a wide spread of verification outcomes and phase realizations; the synthetic subset extends that spread while respecting the gating rules (PID\(\rightarrow\)CSV\(\rightarrow\)DFV/DRC/DCC) and the two\mbox{-}drug design. In the next section we evaluate recent LLMs on both subsets using the phase\mbox{-}aware tasks defined in section \ref{sec:evaluation}.

\section{Evaluation Tasks}
\label{sec:evaluation}
We evaluate postcall extraction. Each model is a general purpose LLM from OpenAI or Google (see §\ref{sec:results}). It takes the final transcript and a task prompt defining the phase schema, gating rules, and JSON format. All runs use temperature 0.
We perform zero-shot inference. We evaluate on two fixed subsets (Section~\ref{sec:dataset-stats}).
We operate on de-identified transcripts (not raw audio); this mirrors common operational auditing workflows that rely on transcripts for evidence retrieval, while leaving multi-modal (audio+text) extensions to future work.

On the real subset we score against human validated reference annotations (§\ref{sec:qa}). On the synthetic subset we score against program validated references produced by the deterministic checker (§\ref{sec:synthetic-data-generation}). We interpret results on the synthetic subset primarily for model ranking and error analysis rather than absolute rates.
We report small, low-latency evaluators in the main tables to match deployment constraints and to avoid confounding on the synthetic subset, where references are program-validated from LLM-generated transcripts. Predicting references derived from a strong model with other strong models risks circularity. We therefore use small models for the primary comparison and include a real-subset sanity check with frontier models.

\paragraph{Research questions:}
(1) \textbf{Phase Boundary Detection:} Can a model mark the start/end turn of each mandated phase with audit\mbox{-}grade precision? \;
(2) \textbf{Compliance Verification:} Given those boundaries, can it decide for each phase whether \emph{Information} (IC) and \emph{Procedural} (PC) conditions hold?\newline
\textbf{Reader note (what is a “phase”?)} A \emph{phase} is a contiguous dialogue segment corresponding to a single auditable checklist item (e.g., “Patient Identification” or “Copay Check”). Phase\mbox{-}aware evaluation is useful because auditors typically ask \emph{where} in a call an obligation was satisfied (and \emph{whether} it was satisfied in the correct order), not only whether the overall call succeeded.
For example, an audit question is often of the form: “Did the assistant verify identity \emph{before} asking about coverage?” Answering this requires locating the relevant span and checking its content, which is exactly what Tasks~1--2 measure.

\textbf{Terminology (for non-specialists).} \emph{IVR} is an automated phone menu (“press 1 for...”); \emph{formulary} indicates whether a drug is covered; \emph{restrictions} refer to prior authorization / step therapy / quantity limits; and \emph{CRN} is the representative name (and optionally a call reference number) recorded for audit logging.

\textbf{Scoring primitives (used by both tasks).}
$S$: a valid span (\verb|turn_range|=\([i,j]\), \(i\le j\)); \;
$A$: phase “ask” flag; \;
$B$: phase “answer” flag; \;
\textbf{NA}: phase is not applicable due to unmet prerequisites.
\emph{Roles:} In \textbf{PID} the \emph{User} (the insurance representative) asks and the \emph{Assistant} answers; in all other scored phases the \emph{Assistant} asks and the \emph{User} answers.
\textbf{NA and gating.} \texttt{"NA"} corresponds to deterministic workflow prerequisites (Section~\ref{sec:annotation}): e.g., if no patient record is found, then coverage verification and downstream drug checks are not applicable and are scored as \texttt{"NA"}.

\emph{Volunteered info} handling in task 2: if \(B{=}\)true but \(A{=}\)false, then IC=true and PC=false for that phase: volunteered information satisfies the information requirement but fails the procedural requirement.
\textbf{Example:} if the representative says “The plan is ACTIVE” before the assistant asks, then coverage status has IC=true but PC=false.
\textbf{Example (PID):} if the assistant states a member ID before being asked, PID has IC=true but PC=false.

\begin{table*}[t]
\centering
\scriptsize
\setlength{\tabcolsep}{5pt}
\renewcommand{\arraystretch}{1.12}
\begin{tabular}{|>{\raggedright\arraybackslash}p{2.6cm}|>{\raggedright\arraybackslash}p{4.1cm}|>{\raggedright\arraybackslash}p{4.0cm}|>{\raggedright\arraybackslash}p{3.8cm}|}
\hline
\textbf{Phase (code)} & \textbf{Ask flag \(A\)} & \textbf{Answer flag \(B\)} & \textbf{Span \(S\)} \\
\hline
IVR (\texttt{ivr}) & — & — & \texttt{ivr.turn\_range} \\
\hline
Greeting (\texttt{greeting}) & — & — & \texttt{greeting.turn\_range} \\
\hline
PID (\texttt{patient\_info}) &
\texttt{patient\_info.\allowbreak user\_asked\_\allowbreak details} &
\texttt{patient\_info.\allowbreak assistant\_provided\_\allowbreak details} &
\texttt{patient\_info.\allowbreak turn\_range} \\
\hline
CSV (\texttt{coverage\_check}) &
\texttt{coverage\_check.\allowbreak assistant\_asked\_\allowbreak status} &
\texttt{coverage\_check.\allowbreak user\_answered\_\allowbreak status} &
\texttt{coverage\_check.\allowbreak turn\_range} \\
\hline
DFV(k) (\texttt{drugs[k].formulary}) &
\texttt{drugs[k].\allowbreak formulary.\allowbreak assistant\_asked} &
\texttt{drugs[k].\allowbreak formulary.\allowbreak user\_answered} &
\texttt{drugs[k].\allowbreak formulary.\allowbreak turn\_range} \\
\hline
DRC(k) (\texttt{drugs[k].restrictions}) &
\texttt{drugs[k].\allowbreak restrictions.\allowbreak assistant\_asked} &
\texttt{drugs[k].\allowbreak restrictions.\allowbreak user\_answered} &
\texttt{drugs[k].\allowbreak restrictions.\allowbreak turn\_range} \\
\hline
DCC(k) (\texttt{drugs[k].copay}) &
\texttt{drugs[k].\allowbreak copay.\allowbreak assistant\_asked} &
\texttt{drugs[k].\allowbreak copay.\allowbreak user\_answered} &
\texttt{drugs[k].\allowbreak copay.\allowbreak turn\_range} \\
\hline
CRN (\texttt{agent\_interaction}) &
\texttt{agent\_interaction.\allowbreak assistant\_asked\_\allowbreak user\_name} &
\texttt{agent\_interaction.\allowbreak user\_name\_\allowbreak provided}\(\neq\texttt{null}\) &
\texttt{agent\_interaction.\allowbreak turn\_range} \\
\hline
\end{tabular}
\caption{Annotation\,$\leftrightarrow$\,phase mapping for the fields used in evaluation. \(A\)=ask, \(B\)=answer, \(S\)=span. Drug phases apply per slot \(k\in\{0,1\}\) (prose: \(d=k+1\)).}
\label{tab:anno_phase_mapping_short}
\end{table*}

\subsection{Task 1: Phase Boundary Detection}
\label{sec:task1}
Given a call transcript \(T=(u_0,\dots,u_{L-1})\) and the phase inventory
\(\mathcal{P}=\{\text{IVR},\text{GRT},\text{PID},\text{CSV},\text{DFV}_{1,2},\text{DRC}_{1,2},\)
\(\text{DCC}_{1,2},\text{CRN}\}\), the model must output, for each phase \(p\), an \emph{inclusive} span \(h_p=[c,d]\), or \texttt{null} if \(p\) does not occur. Reference spans \(g_p=[a,b]\) come from \S\ref{sec:annotation}.\newline
\textbf{Acceptance rules:}
A prediction is correct when it meets the phase\mbox{-}specific rule in Table~\ref{tab:t1_rules_clean}. The small relaxations for IVR and CRN capture common edge cases (repeated IVR prompts; slight drift around the agent name).
Audit settings often require retrieving the precise evidence turns for human review; we therefore treat most phases as exact-span problems, with limited tolerance only where operational ambiguity is common (IVR repetition; short CRN drift).

\begin{table*}[t]
\centering
\scriptsize
\setlength{\tabcolsep}{6pt}
\renewcommand{\arraystretch}{1.1}
\begin{tabular}{|p{5.2cm}|p{8.5cm}|}
\hline
\textbf{Phase} & \textbf{Acceptance rule for predicted span \([c,d]\) vs.\ reference \([a,b]\)} \\
\hline
IVR (\texttt{ivr}) & End equals gold (\(d=b\)); start may begin at or before gold start (\(c\le a\)). \\
\hline
CRN (\texttt{agent\_interaction}) & End equals gold (\(d=b\)); start may differ by at most one turn (\(|c-a|\le 1\)). \\
\hline
GRT, PID, CSV, DFV(k), DRC(k), DCC(k) & Exact match required: \(c=a\) and \(d=b\). \\
\hline
\end{tabular}
\caption{Task~1 (Phase Boundary Detection): span acceptance rules. Drug phases apply per slot \(k\in\{0,1\}\).}
\label{tab:t1_rules_clean}
\end{table*}

\textbf{Metrics and null handling:}
Let \texttt{null} denote “phase absent.”
(i) \textbf{Phase EM (exact match)}: a phase counts as correct if the rule in Table~\ref{tab:t1_rules_clean} holds; \texttt{null} vs.\ \texttt{null} counts as correct; \texttt{null} vs.\ non\mbox{-}\texttt{null} is incorrect.
(ii) \textbf{Turn\mbox{-}level \(F_1\)} is computed on the sets of turns \(P=\{c,\dots,d\}\) and \(G=\{a,\dots,b\}\) only when both spans exist; otherwise the phase is skipped for \(F_1\).
(iii) \textbf{Call\mbox{-}level EM}: fraction of calls in which \emph{all} phases are correct under the rules.
(iv) \textbf{SAD} (\textbf{sum of absolute differences}): positional error when both spans exist, defined as \(\,|a{-}c|+|b{-}d|\,\); skipped otherwise. SAD complements EM by penalizing boundary drift linearly (e.g., a 1-turn miss is less severe than a 10-turn miss).
Intuitively, EM captures strict audit usability, turn-level \(F_1\) captures overlap quality when spans are close but not exact, and SAD quantifies boundary drift.

\subsection{Task 2: Compliance Verification}
\label{sec:task2}
Given the reference spans from Task~1 and the transcript text, the model must decide for each phase whether two conditions hold using \emph{only} explicit evidence:
\emph{Information Compliance (IC)}—the required information was explicitly provided; and
\emph{Procedural Compliance (PC)}—the information was explicitly requested first and then provided.
We use the \(A,B,S\) fields in Table~\ref{tab:anno_phase_mapping_short}. Roles follow the legend; volunteered information is handled uniformly: if \(B\) is true but \(A\) is false, then IC=true and PC=false.\newline

\textbf{Decision rules.}
A phase is \texttt{"NA"} when its preconditions fail (rightmost column of Table~\ref{tab:t2_rules_clean}). Otherwise:
IC is true iff \(B\land S\); PC is true iff \(A\land B\land S\).
(Here, \(S\) denotes that the phase has a valid non-\texttt{null} span; if the span is \texttt{null}, the phase cannot be credited as compliant even if a flag is erroneously set.)
Task~2 isolates reasoning from segmentation: it evaluates compliance decisions given fixed spans (gold or provided spans), allowing us to distinguish boundary errors (Task~1) from compliance-logic errors (Task~2).

\begin{table*}[t]
\centering
\scriptsize
\setlength{\tabcolsep}{6pt}
\renewcommand{\arraystretch}{1.1}
\begin{tabular}{|p{3.1cm}|p{3.7cm}|p{3.7cm}|p{3.5cm}|}
\hline
\textbf{Phase (code)} & \textbf{IC true iff (uses \(B,S\))} & \textbf{PC true iff (uses \(A,B,S\))} & \textbf{When \texttt{"NA"}} \\
\hline
PID (\texttt{patient\_info}) &
\(B \land S\) (assistant provided any identifier; span valid). &
\(A \land B \land S\) (user asked; assistant provided; span valid). &
Never NA due to other phases. \\
\hline
CSV (\texttt{coverage\_check}) &
\(B \land S\) (user stated \texttt{ACTIVE}/\texttt{INACTIVE}). &
\(A \land B \land S\) (assistant asked; user stated status). &
If \texttt{patient\_info.\allowbreak record\_\allowbreak found=\allowbreak false}. \\
\hline
DFV(k) (\texttt{drugs[k].formulary}) &
\(B \land S\) (user said coverage \texttt{YES}/\texttt{NO}; span valid). &
\(A \land B \land S\) (assistant asked; user answered; span valid). &
If CSV was NA or \texttt{coverage \_check.plan\_status}\(\neq\)\texttt{ACTIVE}. \\
\hline
DRC(k) (\texttt{drugs[k].restrictions}) &
\(B \land S\) (user answered PA/ST/QL; span valid). &
\(A \land B \land S\) (assistant asked; user answered; span valid). &
Unless DFV(k) is true and \texttt{drugs[k].formulary.value}=\texttt{YES}. \\
\hline
DCC(k) (\texttt{drugs[k].copay}) &
\(B \land S\) (user stated copay/coinsurance or none; span valid). &
\(A \land B \land S\) (assistant asked; user answered; span valid). &
Unless DRC(k) is true and \texttt{drugs[k].restrictions.value}=\texttt{NO}.\\
\hline
CRN (\texttt{agent\_interaction}) &
IC is always \texttt{"NA"}. &
\(A \land B \land S\) (assistant asked rep name; user provided it). &
IC: always \texttt{"NA"}. \\
\hline
\end{tabular}
\caption{Task~2 (Compliance Verification): IC/PC rules and preconditions. Drug phases apply per slot \(k\in\{0,1\}\). \(A,B,S\) are the fields in Table~\ref{tab:anno_phase_mapping_short}. Volunteered information: \(B{=}\)true, \(A{=}\)false $\Rightarrow$ IC=true, PC=false.}
\label{tab:t2_rules_clean}
\end{table*}

\textbf{Overall fields and metrics:}
\texttt{overall\_IC} is true iff every IC phase with value \(\neq\)\texttt{"NA"} is true; \;
\texttt{overall\_PC} is true iff every PC phase with value \(\neq\)\texttt{"NA"} is true (includes CRN).
We report: (i) \textbf{Phase\mbox{-}level accuracy} (per IC/PC; non\mbox{-}NA phases only); (ii) \textbf{Macro\mbox{-}\(F_1\)} over phases;(iii) \textbf{Per-call Hit}: for each call, the fraction of applicable phases (not \texttt{"NA"}) that are correct; we report the mean \( \pm \) stdev across calls and (iv) \textbf{Call\mbox{-}level accuracy} for \texttt{overall\_IC} and \texttt{overall\_PC}.
Per-call Hit summarizes partial correctness: it is 1.0 only when all applicable phase decisions are correct, and decreases smoothly as more phase decisions are missed.

\textbf{Reproducibility and JSON handling:}
All tasks use fixed, JSON-only prompt templates (Appendix~B) with deterministic decoding (temperature $=0$). In our scoring scripts, outputs that fail JSON parsing or violate the expected key set are treated as invalid predictions for the affected fields (consistent with strict audit-pipeline constraints). We cache model outputs and use standard retries for transient API errors in the release code (Appendix~C).
Appendix~C summarizes the artifact layout and the Task~1/Task~2 driver scripts (including caching and validation guards) used to reproduce the reported metrics.

\section{Results}
\label{sec:results}
We evaluate the models on the two tasks \ref{sec:evaluation} over real and synthetic subsets (Table~\ref{tab:stats}). The synthetic corpus was generated with DeepSeek-family models, but evaluation uses other families (OpenAI GPT-4.x, Google Gemini), reducing circularity and model-family bias. As a sanity check, frontier models on the real subset show the same pattern: modest phase-score gains, but low call-level exact match. Detailed numbers are in the GitHub repo. With only 50 real calls, results are high-quality but statistically limited; the 1,000-call synthetic set gives tighter ranking and stress-test estimates. The repo includes per-phase breakdowns, prompts, and scoring scripts. (Appendix~B--C).
\paragraph{Reading the results - metrics and acronyms:}
For Task~1 we report \textbf{exact match (EM)} of the predicted span under \S\ref{sec:task1} rules, turn-level overlap via \textbf{\(F_1\)}, and \textbf{sum of absolute differences (SAD)} as a boundary-drift measure (lower is better). \textbf{Call-level EM} is the fraction of calls where \emph{all} phases (including correctly predicting \texttt{null} when absent) satisfy the Task~1 acceptance rules.
For Task~2 we report phase-level \textbf{accuracy (Acc)} and \textbf{macro-\(F_1\)} over non-\texttt{"NA"} phases, plus \textbf{Hit}, the mean fraction of applicable phases correct per call (higher is better); \textbf{call-level Acc} corresponds to the overall boolean (\texttt{overall\_IC} or \texttt{overall\_PC}) defined in section \ref{sec:task2}.
We spell out one common acronym here: \textbf{IVR} denotes an \emph{interactive voice response} phone menu (automated prompts); this phase and the greeting often contain repeated prompts and short handoffs that make exact segmentation harder.
\textbf{What Task~2 measures:}
Task~2 evaluates compliance decisions given fixed (reference) spans, so it isolates compliance reasoning from segmentation. In other words, strong Task~2 results should be interpreted as “given correct evidence localization, the model can apply the IC/PC logic reliably,” while Task~1 measures whether models can localize that evidence at audit-grade precision.
\paragraph{Task~1: Phase Boundary Detection:}
Table~\ref{tab:task1_combined_include_null} reports phase-level EM, turn-level \(F_1\), SAD (lower is better), and call-level EM. On real calls, \textbf{Gemini~2.5~Flash} attains the best phase EM (72.7\%) and lowest SAD (23.2), while \textbf{Gemini~2.0~Flash} yields the best turn-level \(F_1\) (0.848). Synthetic results are consistently higher and tighter: \textbf{Gemini~2.0~Flash} leads with EM~80.6\%, \(F_1\)~0.921, and SAD~7.0. Despite strong phase-level scores, \emph{call-level} EM is near-zero on real calls (max 4.0\%) and remains modest on synthetic (max 15.2\%). This gap reflects the multiplicative strictness of exact spans across many phases; single off-by-one errors (often at IVR/GRT or PID/CSV boundaries) collapse call-level EM. Lower latency baselines (GPT-4.1-nano, 4o-mini) show the largest positional drift (e.g., real SAD \(>\) 80), underscoring that accurate segmentation is the principal bottleneck for audit-grade extraction.
\textbf{Why call-level EM collapses.} If a typical call contains on the order of 8--10 scored phases, then even strong phase-level EM can yield low call-level EM under strict conjunction: for example, \(0.73^{10}\approx 0.04\), which matches the best real call-level EM (4.0\%) in Table~\ref{tab:task1_combined_include_null}. This illustrates why small boundary drift (captured by SAD) can still translate into near-zero full-call exactness.

\textbf{Real vs.\ synthetic boundaries:} The real transcripts contain more operational irregularities (e.g., repeated IVR prompts, short confirmations, and noisier handoffs) that make phase transitions less crisp, whereas the synthetic corpus focuses on the audit-relevant dialogue logic and typically omits long hold-time chatter and other frictional artifacts; this contributes to higher and tighter synthetic segmentation scores.

\paragraph{Task~2.1: Information Compliance (IC):}
Per-phase IC is high and stable once spans are fixed (Table~\ref{tab:task2_ic}). On real calls, \textbf{Gemini~2.5~Flash} achieves the top phase accuracy (93.5\%) and call-level accuracy (88.0\%); \textbf{GPT-4.1-mini} posts the highest phase \(F_1\) (0.990) with slightly lower call-level accuracy (80.0\%). Synthetic mirrors these trends: \textbf{Gemini~2.5~Flash} reaches 93.0\% phase accuracy and 80.6\% call-level accuracy. These results indicate that, provided correct boundaries, models reliably judge whether the \emph{required information} was explicitly given (the \(B\land S\) rule), even under distributional shift. Call-level IC accuracy corresponds to the single overall boolean \texttt{overall\_IC} (conjunction over applicable IC phases; \S\ref{sec:task2}); it is therefore sensitive to any single IC error in any applicable phase.

\paragraph{Task~2.2: Procedural Compliance (PC).}
PC adds the \emph{ask-first} requirement (the \(A\land B\land S\) rule). Table~\ref{tab:task2_pc} shows \textbf{Gemini~2.5~Flash} leading on both real (phase accuracy 92.9\%, call-level 96.0\%) and synthetic (91.8\%, 85.5\%) data; \textbf{GPT-4.1-mini} attains the top real phase \(F_1\) (0.867). Variance across model sizes is substantial: smaller models (e.g., GPT-4.1-nano) lag 20–30 points in phase accuracy and 30–40 points in call-level accuracy. Notably, call-level PC is higher on the real subset than on the synthetic subset. This is consistent with corpus composition (Table~\ref{tab:stats}): real calls contain fewer downstream drug phases (e.g., fewer formulary checks), so fewer opportunities exist for PC failure; conversely, synthetic calls deliberately exercise the full gate structure, increasing the number of required PC decisions per call. IC and PC call-level accuracies are computed over different overall fields (\texttt{overall\_IC} vs.\ \texttt{overall\_PC}) with different applicability patterns (e.g., CRN contributes to \texttt{overall\_PC} but IC for CRN is always \texttt{"NA"}). Thus, IC and PC call-level percentages are not directly comparable as “harder vs.\ easier” across tasks; they should be interpreted within-task.\newline
\textbf{Takeaways:}
\textit{(1) Boundary detection is the bottleneck} -- models can score high per phase (F1 \(0.84\)--\(0.92\)) but call-level EM on real calls is $\leq$4\%\ because a single boundary slip fails the call. \textit{(2) With fixed spans, compliance is strong} -- top models reach >93\%\ phase accuracy and 88–96\%\ call-level IC/PC, while mid-tier models trail by 10–20 points. \textit{(3) Synthetic data preserves ranking} -- ordering is broadly consistent across real and synthetic sets, making synthetic calls useful for stress tests and forecasting real trends. \textit{(4) Errors concentrate in volunteered-info cases, A/B role flips, and gating transitions (DFV→DRC→DCC)}, which small prompt tweaks or rule checks can reduce.
\begin{table}[t]
\caption{Task 1 (Phase Boundary Detection) Main results use small, low-latency models on 50 real and 1,000 synthetic conversations; frontier real-subset evaluations are included in the submission package.}
\label{tab:task1_combined_include_null}
\resizebox{\columnwidth}{!}{
\begin{tabular}{llcccccc}
\toprule
& & \multicolumn{3}{c}{Phase-level} & & \multicolumn{1}{c}{Call-level} \\
\cmidrule{3-5} \cmidrule{7-7}
Model & Dataset & EM (\%) ↑ & F1 ↑ & SAD ↓ & & EM (\%) ↑ \\
\midrule
\multirow{2}{*}{GPT-4o-mini}
& Real & 53.6 & 0.661 & 89.4 & & 0.0 \\
& Synthetic & 62.6 & 0.813 & 25.6 & & 0.1 \\
\midrule
\multirow{2}{*}{GPT-4.1-mini}
& Real & 63.6 & 0.766 & 51.1 & & 0.0 \\
& Synthetic & 73.6 & 0.891 & 9.2 & & 4.8 \\
\midrule
\multirow{2}{*}{GPT-4.1-nano}
& Real & 40.9 & 0.537 & 102.7 & & 0.0 \\
& Synthetic & 56.5 & 0.729 & 41.5 & & 0.5 \\
\midrule
\multirow{2}{*}{Gemini 2.5 Flash}
& Real & \textbf{72.7} & 0.842 & \textbf{23.2} & & \textbf{4.0} \\
& Synthetic & 79.7 & 0.911 & 7.4 & & 8.1 \\
\midrule
\multirow{2}{*}{Gemini 2.0 Flash}
& Real & 70.2 & \textbf{0.848} & 34.7 & & 0.0 \\
& Synthetic & \textbf{80.6} & \textbf{0.921} & \textbf{7.0} & & \textbf{15.2} \\
\midrule
\multirow{2}{*}{Gemini 2.5 Flash Lite}
& Real & 59.5 & 0.710 & 63.0 & & 0.0 \\
& Synthetic & 73.1 & 0.867 & 12.6 & & 4.1 \\
\bottomrule
\end{tabular}
}
\end{table}
\begin{table}[t]
\caption{Evaluation on task 2.1 -- Information Compliance.}
\label{tab:task2_ic}
\resizebox{\columnwidth}{!}{
\begin{tabular}{llcccccc}
\toprule
& & \multicolumn{3}{c}{Phase-level} & & \multicolumn{1}{c}{Call-level} \\
\cmidrule{3-5} \cmidrule{7-7}
Model & Dataset & Acc (\%) ↑ & Hit (\%) ↑ & F1 ↑ & & Acc (\%) ↑ \\
\midrule
\multirow{2}{*}{GPT-4o-mini}
& Real & 84.5 & 78.0 & 0.719 & & 26.0 \\
& Synthetic & 83.0 & 77.5 & 0.809 & & 34.2 \\
\midrule
\multirow{2}{*}{GPT-4.1-mini}
& Real & 90.8 & 89.6 & \textbf{0.990} & & 80.0 \\
& Synthetic & 86.6 & 83.2 & 0.925 & & 56.1 \\
\midrule
\multirow{2}{*}{GPT-4.1-nano}
& Real & 79.2 & 75.6 & 0.822 & & 46.0 \\
& Synthetic & 70.0 & 66.6 & 0.802 & & 39.5 \\
\midrule
\multirow{2}{*}{Gemini 2.5 Flash}
& Real & \textbf{93.5} & \textbf{92.9} & 0.978 & & \textbf{88.0} \\
& Synthetic & \textbf{93.0} & \textbf{91.6} & \textbf{0.971} & & \textbf{80.6} \\
\midrule
\multirow{2}{*}{Gemini 2.0 Flash}
& Real & 90.0 & 88.2 & 0.932 & & 74.0 \\
& Synthetic & 85.9 & 81.9 & 0.948 & & 49.6 \\
\midrule
\multirow{2}{*}{Gemini 2.5 Flash Lite}
& Real & 83.8 & 76.9 & 0.765 & & 22.0 \\
& Synthetic & 81.2 & 76.1 & 0.852 & & 36.1 \\
\bottomrule
\end{tabular}
}
\end{table}
\begin{table}[t]
\caption{Evaluation on task 2 -- Procedural Compliance.}
\label{tab:task2_pc}
\resizebox{\columnwidth}{!}{
\begin{tabular}{llcccccc}
\toprule
& & \multicolumn{3}{c}{Phase-level} & & \multicolumn{1}{c}{Call-level} \\
\cmidrule{3-5} \cmidrule{7-7}
Model & Dataset & Acc (\%) ↑ & Hit (\%) ↑ & F1 ↑ & & Acc (\%) ↑ \\
\midrule
\multirow{2}{*}{GPT-4o-mini}
& Real & 80.7 & 79.0 & 0.637 & & 64.0 \\
& Synthetic & 80.2 & 77.4 & 0.772 & & 52.9 \\
\midrule
\multirow{2}{*}{GPT-4.1-mini}
& Real & 88.7 & 88.0 & \textbf{0.867} & & 82.0 \\
& Synthetic & 84.7 & 81.6 & 0.904 & & 54.2 \\
\midrule
\multirow{2}{*}{GPT-4.1-nano}
& Real & 68.0 & 66.6 & 0.758 & & 54.0 \\
& Synthetic & 58.9 & 57.3 & 0.785 & & 43.0 \\
\midrule
\multirow{2}{*}{Gemini 2.5 Flash}
& Real & \textbf{92.9} & \textbf{93.2} & 0.820 & & \textbf{96.0} \\
& Synthetic & \textbf{91.8} & \textbf{91.2} & \textbf{0.951} & & \textbf{85.5} \\
\midrule
\multirow{2}{*}{Gemini 2.0 Flash}
& Real & 84.9 & 83.6 & 0.811 & & 72.0 \\
& Synthetic & 80.0 & 78.1 & 0.855 & & 60.6 \\
\midrule
\multirow{2}{*}{Gemini 2.5 Flash Lite}
& Real & 73.1 & 73.0 & 0.226 & & 72.0 \\
& Synthetic & 69.8 & 69.4 & 0.675 & & 66.4 \\
\bottomrule
\end{tabular}
}
\end{table}
\newline \textbf{Summary:} The results validate the design of INSURE\mbox{-}Dial: segmentation (Task~1) is the critical failure point for end-to-end auditability, whereas compliance classification (Task~2) becomes reliable for top-tier models once spans are correct. Synthetic data preserves ranking and reveals stress points (e.g., more required drug phases), while real calls expose realistic IVR/greeting drift that challenges exact spans.
Frontier real-subset checks (Gemini 2.5 Pro, GPT-4.1, o3, Claude Sonnet 4.0) modestly increase phase EM and F1 but do not close the call-level EM gap; see repository.
\section{Related Work}
\paragraph{Task‑oriented dialogue (TOD) and turn‑level evaluation:}
Most TOD corpora evaluate utterances for intent/slot accuracy rather than end‑to‑end auditability. Landmark resources (e.g., MultiWOZ, SGD, STAR, Taskmaster, ToolDial, DialogTool) and their leaderboards emphasize local turn quality and dialogue success proxies~\cite{budzianowski2018multiwoz,rastogi2020towards,mosig2020star,sidahmed2024parameter,shimtooldial,wang-etal-2025-rethinking-stateful}.
Long‑context stress tests show that strong turn‑level scores do not guarantee trajectory reliability as conversations grow: agents drop context and fail goals despite fluent local turns~\cite{liu2023agentbench,sirdeshmukh2025multichallenge,krishna2023longeval,modarressi2024ret}.
Agent‑centric benchmarks extend this point by testing tool use, planning, and long‑horizon control~\cite{liu2023agentbench}; complementary long‑text evaluations report degradation with increasing context length and structure~\cite{longeval2025}. 
Industry studies comparing long‑context vs.\ retrieval also demonstrate design trade‑offs that are orthogonal to compliance~\cite{li-etal-2024-retrieval}. Segment-quality metrics (e.g., DynaEval, Breakdowns, FineD) move beyond single turns, but their spans target cohesion or coherence rather than statutory checkpoints~\cite{zhang2021dynaeval,braggaar-2023-breakdowns,zhang2022fined}. In parallel, our recent OIP--SCE formalism frames regulated dialogue auditing as \emph{order-aware progress through auditable phases} with explicit evidence requirements, motivating phase-boundary and phase-level compliance evaluation beyond generic TOD success metrics~\cite{kulkarni2026all}.
\paragraph{Voice agents and call-center datasets:}
Recent releases focus on real or simulated contact-center conversations but stop short of regulation-grade, phase-aligned labeling. \emph{Dial-In-LLM}~\cite{hong2024dial} contributes $\sim$100k real bank customer-service calls (Mandarin), transcribed and anonymized for downstream tasks, yet without the ordered compliance phases our audits require. \emph{Automated Survey Collection with LLM-based Conversational Agents}~\cite{kaiyrbekov2025automated} demonstrates end-to-end AI telephone surveys in 2025, again lacking IVR/identity/coverage/copay labels needed for HIPAA/CMS-style verification. \emph{CallCenterEN}~\cite{dao2025real} offers 92k English call-center transcripts/scripts with useful metadata but no phase-gated compliance fields. Collectively, these works advance voice-agent data and tooling, but they do not provide the \emph{structured}, span-level, \emph{ordered} phase annotations (e.g., IVR $\rightarrow$ identity $\rightarrow$ coverage $\rightarrow$ drug checks $\rightarrow$ representative ID) needed to verify that every mandated step occurred in sequence under workflow gating. Our prior OIP--SCE work formalizes this phase- and order-aware audit view~\cite{kulkarni2026all}; INSURE-Dial fills the remaining resource gap by releasing a public benchmark with phase spans, ask/answer roles, and deterministic preconditions tailored to insurance benefit verification.

\paragraph{Safety, policy adherence, and red‑teaming:}
Safety suites probe for disallowed content or robustness under attack, an important but different axis from regulatory auditability. Recent work includes HarmBench~\cite{mazeika2024harmbench}, GuardBench~\cite{guardbench2024}, SALAD‑Bench~\cite{li2024salad}, and multi‑round jailbreak studies documenting interaction‑level vulnerabilities~\cite{zhou2024multi}. Automated red‑teaming has advanced rapidly (e.g., APRT and multilingual multi‑turn variants), highlighting the need for standardized evaluation protocols~\cite{jiang2025automated,singhania2025multi}. These contributions surface risk hot‑spots but do not certify that every legally required conversational phase was explicitly completed in the correct order.
\paragraph{Compliance-aware evaluation.}
Prior NLP work on privacy/factuality audits (e.g., DialFact, privacy-policy parsing, KYC-style checks) typically targets specific obligations or reply correctness rather than workflow-level completion across \emph{ordered} phases~\cite{gupta-etal-2022-dialfact,fan2024goldcoin}.
Our recent OIP--SCE work proposes a general, order-aware view of conversational auditing in regulated settings by representing each mandated checklist item as an \emph{auditable phase} with explicit evidence and sequencing constraints~\cite{kulkarni2026all}.
By contrast, \textbf{INSURE-Dial} contributes the first public \emph{benchmark dataset and tasks} (to our knowledge) that instantiate this phase-level audit view for insurance benefit verification calls: each call is segmented into contiguous, regulation-salient phases with explicit ask/answer roles and preconditions, enabling two tasks—\emph{Phase Boundary Detection} and \emph{Compliance Verification}—that directly test the unit auditors actually score. This moves assessment beyond turn-local quality or single-violation safety toward auditable, end-to-end reliability in regulated voice automation.

\section{Conclusion}
\label{sec:conclusion}
We present \textbf{INSURE\mbox{-}Dial}, a phase\mbox{-}aware benchmark and schema for auditing insurance benefit\mbox{-}verification calls. The corpus (50 real, 1{,}000 synthetic) is annotated with explicit spans and \(A/B\) flags, enabling two targeted evaluations: \emph{Phase Boundary Detection} and \emph{Compliance Verification}. Experiments with contemporary LLMs show that: (i) exact span recovery is the bottleneck that suppresses call-level success; (ii) with spans fixed, both information and procedural compliance can be judged accurately under deterministic rules; and (iii) synthetic calls can broaden coverage while preserving real-world model ranking and error patterns. De-identified data are included in an anonymized review package. Future work: align acceptance rules with auditing tolerances, add active learning for boundary correction, and extend to adjacent domains (finance, clinical). Ultimately, evaluation must match what regulators audit: every required phase occurred, in order, with explicit asks and answers. INSURE\mbox{-}Dial operationalizes it.

\section*{Limitations}
INSURE\mbox{-}Dial captures one workflow in U.S. pharmacy benefit verification, in English, with at most two drug checks per call. Generalization to other jurisdictions, languages, inbound calling, or different healthcare tasks is not established. The real subset is small (50 calls). The synthetic subset is larger but program-validated rather than fully human-validated; we therefore use it mainly for model ranking and error analysis rather than absolute rates. Transcripts are produced by ASR, so span labels may reflect ASR errors; raw audio is not released, which limits independent re-segmentation. Phase acceptance rules are intentionally strict; alternative tolerances could raise call-level exact match or change rankings. Evaluation depends on vendor LLM APIs with temperature set to 0; API drift and model updates can change numbers despite cached outputs. Synthetic generation uses LLMs and may inherit topical or stylistic biases. The corpus is rebalanced to exercise rare outcomes, so its distribution may differ from live traffic. Finally, our ask/answer policy and treatment of volunteered information reflect one operational interpretation; other organizations may define these differently.

\section*{Ethical Considerations}
\label{sec:ethics}

\paragraph{Data source and consent:}
The real-call portion of INSURE-Dial consists of business-to-business (B2B) insurance-benefits-verification calls made on behalf of a partner health system to health insurance companies as part of routine operations. For the public release, the partner health system reviewed the 50 specific, de-identified transcripts to ensure no residual privacy risks existed before authorizing the release for research/public-interest purposes. Standard operational notifications (e.g., ``this call may be recorded for quality assurance'') were also in place for the insurance representatives to hear and acknowledge.

\paragraph{Modality choice (text only; no audio):}
We release \emph{only} de-identified text transcripts and structured JSON annotations; we do not release raw audio. This modality choice is privacy-motivated: releasing text transcripts enables HIPAA Safe Harbor de-identification, whereas releasing audio would make public release infeasible. This choice is also aligned with current operational auditing practice, which primarily relies on transcripts. We view INSURE-Dial as a text-logic backbone on which future work (including multimodal extensions) can build without increasing release risk.

\paragraph{De-identification and privacy protections:}
We follow HIPAA Safe Harbor (45 CFR \S164.514(b)(2)): direct identifiers (names, phone/email, precise dates, IDs) are removed or masked prior to any annotation or release. Automated scrubbing is followed by human verification performed independently by staff at the partner health system. Data were handled on encrypted, access-controlled systems with limited access, and annotators were trained internal staff.

\paragraph{Unreleased operational data:}
Beyond the released subset, we collected a much larger set of operational calls. The remaining $\sim$19{,}950 calls are not part of the release package and were not human-annotated due to resource constraints. They remain stored in encrypted, access-controlled internal systems. These calls were used only to characterize the broader data distribution (e.g., seeding parameters for the synthetic generator) and are not released.

\paragraph{Human annotators and quality assurance:}
Human validation was performed by trained internal annotators (not crowdworkers). Two trained annotators independently labeled the real calls, with disagreements adjudicated by a senior reviewer. We report high inter-annotator agreement on the double-annotated subset (Cohen's $\kappa=0.95\pm0.02$). Recruitment, payment, and demographic characteristics are not disclosed; annotators were internal staff. As the data are fully de-identified operational transcripts, we did not require IRB approval for release.

\paragraph{Synthetic data and safeguards:}
The synthetic portion (1{,}000 calls) contains no real-world patient or policy-holder information. Synthetic calls are generated to mirror the same phase structure and outcome diversity as the real workflow, under the same phase schema and gating rules, and are screened by a deterministic program validator. Any generation failing schema or gating logic is rejected and regenerated, guaranteeing structural validity and label consistency for the released synthetic subset. In practice, approximately $\sim$76\% of raw generations pass the deterministic validator on the first attempt, with the remainder rejected or revised until they satisfy all rules. In addition, multiple trained annotators manually reviewed $\sim$100 synthetic calls; roughly $\sim$6\% showed minor stylistic issues and none exhibited hard schema violations (e.g., missing mandatory phases or impossible outcome combinations).

\paragraph{Potential risks and intended use:}
The primary risks are privacy breaches and out-of-scope application. We mitigate privacy risk by releasing text-only transcripts (no audio), enforcing HIPAA Safe Harbor de-identification with manual verification, and limiting release to a small, reviewed subset of real calls. The dataset and related code released as part of this paper have a permissible MIT license. The dataset reflects U.S.\ pharmacy benefit verification; applying it elsewhere may require adapting phase definitions and auditing logic.

\paragraph{Known limitations affecting downstream use:}
Automatic speech recognition can introduce transcription errors, and synthetic data may differ from live traffic. In particular, real calls contain operational friction (e.g., hold times, transfers, connection checks, repeated prompts) that can inflate turn counts, while synthetic generation intentionally targets the audit-relevant dialogue logic and omits long holds or repetitive connectivity issues. We document these factors and report real and synthetic results separately.

\paragraph{Use of AI assistants in writing:}
We used AI assistants minimally for grammar correction and phrasing improvements in some sections of the paper. No content, experiments, or analysis were generated by AI tools.

\paragraph{Goal:}
Our goal is to support research on transparent, auditable dialogue systems and to provide a benchmark that measures the gap between current model behavior and audit requirements, while discouraging uses that violate privacy or regulatory obligations.

\bibliography{custom}
\appendix
\section{Phase Flow, Audit Schema, and Annotation Guidelines}
\label{app:annotation}

\begin{figure*}[t]
  \centering
  \includegraphics[page=1,width=\textwidth]{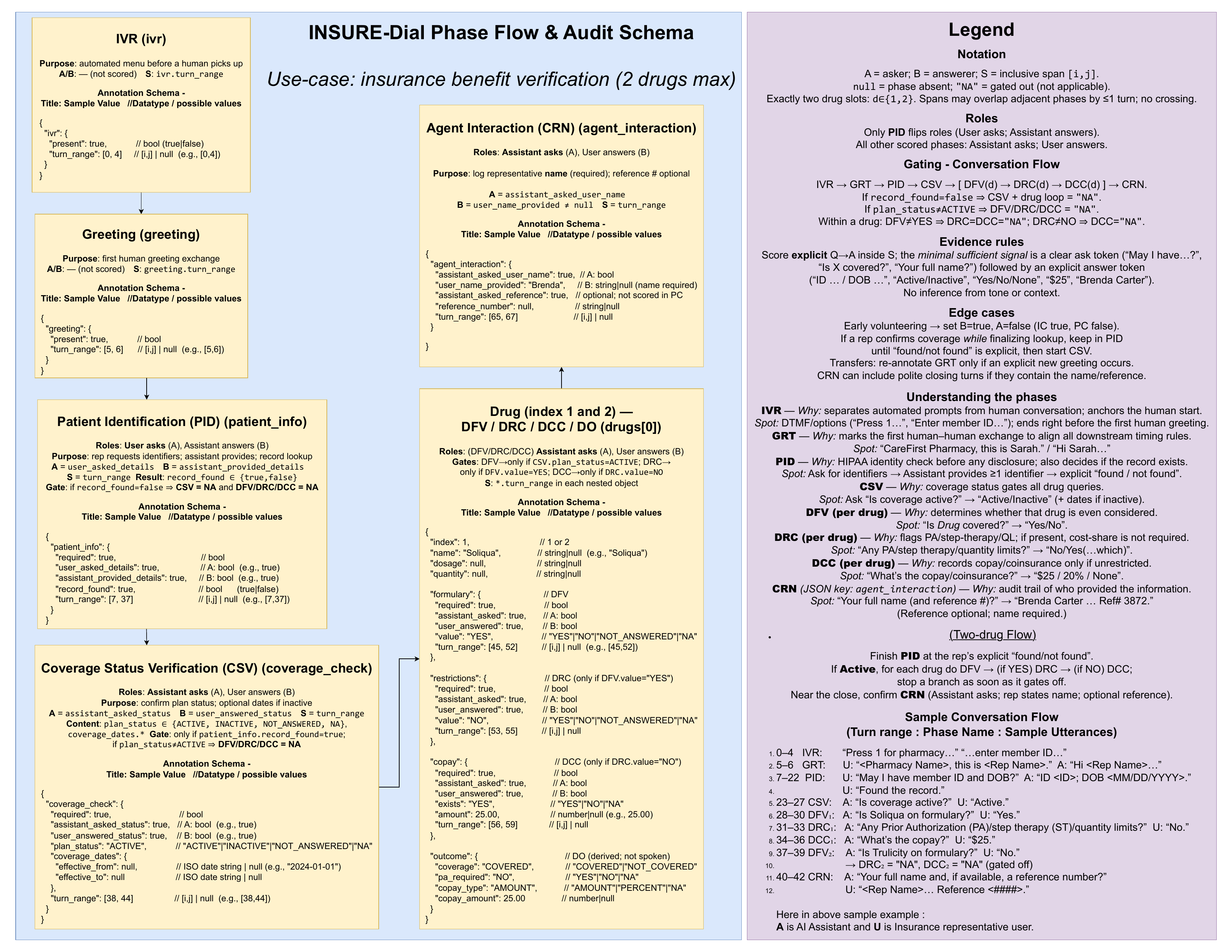}
  \caption{INSURE-Dial phase flow and audit schema. Each phase stores role-typed ask/answer flags ($A,B$), an inclusive span $S=[i,j]$, and minimal content fields; arrows show gating (PID$\rightarrow$CSV$\rightarrow$DFV/DRC/DCC per drug slot, then CRN).}
  \label{fig:app_framework}
\end{figure*}

\subsection{Core definitions}
\label{app:core-defs}
\noindent\textbf{Turns and spans:}
A \emph{turn} is a single utterance by either the AI assistant (Patient Care Specialist) or the insurance representative.
Turns are indexed from 0. A \emph{turn range} (span) is inclusive: $S=[i,j]$ denotes turns $i,i{+}1,\dots,j$.

\noindent\textbf{Absent vs.\ gated-out:}
We distinguish (i) \texttt{null} spans for phases that do not occur, from (ii) \texttt{"NA"} values for phases that are not applicable due to gating/prerequisites (see \S\ref{app:gating}).

\noindent\textbf{Ask/answer flags:}
Each scored phase (except IVR/GRT) includes:
$A$ = an explicit request for the target information in that phase; \;
$B$ = an explicit provision of that information; \;
$S$ = a valid, non-\texttt{null} span. \;
Volunteered information is allowed: $B{=}\texttt{true}$ with $A{=}\texttt{false}$.

\subsection{Roles and phase inventory}
\label{app:phases}
\noindent\textbf{Role convention.}
Only \textbf{PID} flips roles: the \emph{insurance representative} requests identifiers (User asks), and the \emph{assistant} provides them (Assistant answers).
In all other scored phases, the assistant asks and the representative answers.

\vspace{4pt}
\noindent\textbf{Phases (ordered).}
We annotate calls using the following checklist-aligned phases.

\begin{itemize}\setlength{\itemsep}{2pt}
  \item \textbf{IVR} (\texttt{ivr}): automated menu before a human answers; span only (no $A/B$). Keys: \texttt{present}, \texttt{turn\_range}.
  \item \textbf{Greeting} (\texttt{greeting}): first assistant--representative greeting exchange; span only. Keys: \texttt{present}, \texttt{turn\_range}.
  \item \textbf{Patient Identification} (\texttt{patient\_info}; PID): representative requests identifiers; assistant provides them; representative confirms whether the record is found. Keys: \texttt{user\_asked\_details} ($A$), \texttt{assistant\_provided\_details} ($B$), \texttt{record\_found}, \texttt{turn\_range}.
  \item \textbf{Coverage Status} (\texttt{coverage\_check}; CSV): assistant asks whether the plan is active; representative answers. Keys: \texttt{assistant\_asked\_status} ($A$), \texttt{user\_answered\_status} ($B$), \texttt{plan\_status} $\in$ \{\texttt{ACTIVE}, \texttt{INACTIVE}, \texttt{NOT\_ANSWERED}, \texttt{NA}\}, optional \texttt{coverage\_dates}, \texttt{turn\_range}.
  \item \textbf{Drug loop} (\texttt{drugs[k]}; $k\in\{0,1\}$): exactly two drug slots are allocated (drug slot 1 = \texttt{drugs[0]}, drug slot 2 = \texttt{drugs[1]}). Each slot contains:
    \begin{itemize}\setlength{\itemsep}{1pt}
      \item \textbf{Formulary} (\texttt{formulary}; DFV): is the drug covered / on formulary? \texttt{value} $\in$ \{\texttt{YES}, \texttt{NO}, \texttt{NOT\_ANSWERED}, \texttt{NA}\}.
      \item \textbf{Restrictions} (\texttt{restrictions}; DRC): are restrictions present (PA/ST/QL)? \texttt{value} $\in$ \{\texttt{YES}, \texttt{NO}, \texttt{NOT\_ANSWERED}, \texttt{NA}\}.
      \item \textbf{Copay} (\texttt{copay}; DCC): copay/coinsurance amount or none. \texttt{exists} $\in$ \{\texttt{YES}, \texttt{NO}, \texttt{NA}\}; \texttt{amount} is numeric if explicitly stated.
      \item \textbf{Outcome} (\texttt{outcome}): derived fields (not spoken) used for consistency checks (e.g., covered vs.\ not covered; copay type/amount).
    \end{itemize}
    Each DFV/DRC/DCC sub-phase stores \texttt{required}, $A$, $B$, content value(s), and \texttt{turn\_range}.
  \item \textbf{Agent Interaction} (\texttt{agent\_interaction}; CRN): assistant requests the representative’s full name (required) and optionally a call reference number; used for audit logging. Keys: \texttt{assistant\_asked\_user\_name} ($A$), \texttt{user\_name\_provided} ($B$, stored as string; name is required for $B{=}\texttt{true}$), optional \texttt{assistant\_asked\_reference}, optional \texttt{reference\_number}, \texttt{turn\_range}.
\end{itemize}

\subsection{Conversation flow and gating rules}
\label{app:gating}
\noindent\textbf{Canonical flow:}
\[
\resizebox{\columnwidth}{!}{$
\text{IVR}\rightarrow \text{GRT}\rightarrow \text{PID}\rightarrow \text{CSV}\rightarrow
[\text{DFV}(k)\rightarrow \text{DRC}(k)\rightarrow \text{DCC}(k)]_{k\in\{0,1\}}\rightarrow \text{CRN}
$}
\]

\noindent\textbf{Gating (preconditions).}
We mark downstream phases \texttt{"NA"} when prerequisites fail:
\begin{enumerate}\setlength{\itemsep}{2pt}
  \item If \texttt{patient\_info.record\_found=false}, then \textbf{CSV} and the entire drug loop are \texttt{"NA"}.
  \item If \texttt{coverage\_check.plan\_status != "ACTIVE"}, then \textbf{DFV/DRC/DCC} for both drugs are \texttt{"NA"}.
  \item Within each drug slot $k\in\{0,1\}$:
    \begin{itemize}\setlength{\itemsep}{1pt}
      \item If \texttt{formulary.value != YES}, then \textbf{DRC} and \textbf{DCC} are \texttt{"NA"}.
      \item If \texttt{restrictions.value != NO}, then \textbf{DCC} is \texttt{"NA"}.
    \end{itemize}
\end{enumerate}

\subsection{Evidence rules}
\label{app:evidence}
We annotate and score \emph{only explicit evidence} inside the phase span $S$.
A minimal sufficient signal for $A$ is an explicit request token (e.g., “May I have…?”, “Is $x$ covered?”, “Your full name?”).
A minimal sufficient signal for $B$ is an explicit answer token (e.g., an ID/DOB, \texttt{ACTIVE}/\texttt{INACTIVE}, \texttt{YES}/\texttt{NO}, a dollar amount, a representative name).
We do not infer intent from tone, context, or unstated implications.

\subsection{Edge cases and boundary conventions}
\label{app:edgecases}
\begin{itemize}\setlength{\itemsep}{2pt}
  \item \textbf{Early volunteering:} if the representative volunteers information before an explicit request, annotate $B{=}\texttt{true}$ and $A{=}\texttt{false}$ (information satisfied; procedure not).
  \item \textbf{PID--CSV overlap:} if the representative confirms coverage while still finalizing record lookup, keep the turn(s) in PID until the rep explicitly states “found/not found.”
  \item \textbf{Transfers:} re-annotate \textbf{Greeting} only if an explicit new greeting exchange occurs; otherwise continue the existing phase progression.
  \item \textbf{Farewells:} exclude pure closings from spans; include closing turns in CRN only if they contain the representative name and/or reference number.
\end{itemize}

\subsection{Quality assurance checklist (for human validation and programmatic checks)}
\label{app:qa}
We enforce the following invariants prior to scoring:
\begin{enumerate}\setlength{\itemsep}{2pt}
  \item \textbf{Span validity:} all non-\texttt{null} spans are inclusive \texttt{[i,j]} with $0\le i\le j\le L{-}1$.
  \item \textbf{No crossings; limited overlap:} spans do not cross; adjacent phases may overlap by at most one turn.
  \item \textbf{Gating correctness:} \S\ref{app:gating} is enforced exactly; gated-out phases use \texttt{"NA"}.
  \item \textbf{Role consistency:} PID uses (User asks, Assistant answers); all other scored phases use (Assistant asks, User answers).
  \item \textbf{Two-drug shape:} exactly two \texttt{drugs[*]} entries; if one slot is unused, it is fully filled with \texttt{null}/\texttt{"NA"} as appropriate.
  \item \textbf{Value constraints:} enums match spec; numeric fields are non-negative; derived \texttt{outcome} is consistent with DFV/DRC/DCC values.
\end{enumerate}

\subsection{Illustrative example}
\label{app:example}
\vspace{2pt}
\noindent\textbf{Example trace (phase cues).}
IVR: “Press 1 for pharmacy…” $\rightarrow$
GRT: “Thank you for calling…, this is …” $\rightarrow$
PID: rep requests member ID/DOB; assistant provides; rep says “Found the record” $\rightarrow$
CSV: assistant asks “Is coverage active?”; rep answers \texttt{ACTIVE} or \texttt{INACTIVE} $\rightarrow$
DFV/DRC/DCC per drug (gated as needed) $\rightarrow$
CRN: assistant asks rep name (and optional reference); rep provides.

\section{Prompts and Templates}
\label{app:prompts}

This appendix provides the exact prompt templates used in our benchmark pipeline, per reviewer request. Unless noted otherwise, all runs use temperature 0 and require models to return well-formed JSON only.

\subsection{Task 1 Prompt: Phase Boundary Detection}
\label{app:prompt_task1}
\noindent\textbf{Purpose.} Given a transcript, the model outputs an \emph{inclusive} turn span for each phase (or \texttt{null} if absent), following the phase inventory and formatting constraints described in \ref{sec:evaluation}.

Prompt:-
\begin{PromptBlock}
You are an expert annotator tasked with labeling healthcare coverage verification phone calls between an AI Patient Care Specialist ("Assistant") and a Human Insurance Representative ("User"). Produce ONLY the specified JSON schema—no additional commentary.

### JSON Output Schema:

For each field below, provide either:

* `[start_turn, end_turn]` (inclusive integers)
* `null` (if the described event does not occur)

```json schema
{
  "ivr": [int, int] | null,
  "greeting": [int, int] | null,
  "patient_info": [int, int] | null,
  "coverage_check": [int, int] | null,
  "drug1_formulary": [int, int] | null,
  "drug1_restrictions": [int, int] | null,
  "drug1_copay": [int, int] | null,
  "drug2_formulary": [int, int] | null,
  "drug2_restrictions": [int, int] | null,
  "drug2_copay": [int, int] | null,
  "agent_interaction": [int, int] | null
}
```

---

### Annotation Guidelines (Follow Strictly)

#### General Definitions:

* **Turn**: Each utterance by either Assistant or User, indexed from 0.
* **Turn Range**: Inclusive `[start_turn, end_turn]` for related utterances.
* **Null**: Use explicitly if event never occurs.
* **No overlaps**: Consecutive phases must overlap by at most one turn.

#### Phase-specific Guidelines:

### 1. IVR (Interactive Voice Response)

* IVR spans start from the first automated speech (structured instructions, DTMF options like "Press 1", "Enter ID") and end immediately before the human representative's first utterance (typically a greeting like "Thank you for calling...").
* Set as `null` if IVR is never present.
* Start from turn 0

### 2. Greeting

* Span starts from initial explicit human greeting ("Hello", "Hi", etc.) between Assistant/User.
* Ends once greeting exchange concludes.
* Set as `null` if no explicit greeting occurs.

### 3. Patient Information

* Begins immediately after greeting ends.
* Ends explicitly after User confirms patient record found or not found.
* Includes troubleshooting, verification, or repeated attempts.
* Set as `null` only if never explicitly initiated.

### 4. Coverage Check

* Annotated only if patient record explicitly confirmed as found.
* Starts from Assistant’s first explicit question on coverage status.
* Ends with User’s explicit confirmation (ACTIVE, INACTIVE, NOT ANSWERED).
* Set as `null` if patient record not found.

### 5-10. Drug Information (Exactly two drugs)

* **Formulary Check:** Assistant explicitly asks if drug is covered.
* **Restrictions Check:** Annotate only if formulary = YES. Assistant explicitly asks about prior authorization (PA), step therapy, or quantity limits.
* **Copay Check:** Annotate only if restrictions explicitly = NO. Assistant explicitly asks about copay or coinsurance amounts.
* Set as `null` if phase condition not met (e.g., formulary NO → restrictions/copay null).
* If fewer than two drugs discussed, remaining drug fields are `null`.

### 11. Agent Interaction

* Span begins from Assistant's explicit request for User’s full name or call/reference number.
* Ends at the last turn explicitly providing User’s name or reference number.
* **Include:** All turns explicitly requesting/providing name/reference.
* **Exclude:** Generic thank-yous, goodbyes, small talk, early volunteering before Assistant's explicit request, and farewell-only turns.
* If final metadata exchange combined with polite closing, include entire turn.

---

### Critical Edge Case Instructions:

* **Early volunteering:** If either speaker provides phase-relevant information before the usual explicit request, do not ignore it. Include those turns in the appropriate phase span if they clearly contain the target evidence for that phase (e.g., plan status, formulary yes/no, restrictions, copay, rep name). Task 1 is span-only; it should localize where the relevant evidence occurs, regardless of whether it was requested first.
* **Multiple farewells:** Exclude farewell-only turns.
* **Transferred calls:** Only annotate greeting if explicitly exchanged again.

---

### Final Checklist Before Submission:

* Exactly 11 fields, adhering strictly to above guidelines.
* Maximum one-turn overlap between consecutive phases.
* Clearly mark `null` for events not occurring.

---

Use the provided guidelines precisely to ensure accurate and consistent annotations.

## Transcript (array of dicts with role, text, turn):

## {transcript_here}
\end{PromptBlock}

\subsection{Task 2 Prompt: Compliance Verification}
\label{app:prompt_task2}
\noindent\textbf{Purpose.} Given the transcript and \emph{fixed} phase spans, the model predicts phase-level compliance labels under the deterministic IC/PC definitions in \ref{sec:task2}. Outputs must be JSON with exactly the required keys and label set.

Prompt:-
\begin{PromptBlock}
You are a compliance specialist evaluating a healthcare-coverage verification call between an AI Patient-Care Specialist (“Assistant”) and a Human Insurance Representative (“User”).

Return **one—and only one—JSON object** in the exact form shown below.
Use the literals **true**, **false**, or the quoted string **"NA"**.
Do **not** add comments, extra keys, or trailing commas.

–––––  OUTPUT SCHEMA  –––––

{
  "Information_Compliance": {
    "PID": true | false | "NA",
    "CSV": true | false | "NA",
    "DFV_drug_1": true | false | "NA",
    "DRC_drug_1": true | false | "NA",
    "DCC_drug_1": true | false | "NA",
    "DFV_drug_2": true | false | "NA",
    "DRC_drug_2": true | false | "NA",
    "DCC_drug_2": true | false | "NA",
    "overall_IC": true | false
  },
  "Procedural_Compliance": {
    "PID": true | false | "NA",
    "CSV": true | false | "NA",
    "DFV_drug_1": true | false | "NA",
    "DRC_drug_1": true | false | "NA",
    "DCC_drug_1": true | false | "NA",
    "DFV_drug_2": true | false | "NA",
    "DRC_drug_2": true | false | "NA",
    "DCC_drug_2": true | false | "NA",
    "CRN": true | false | "NA",
    "overall_PC": true | false
  }
}

–––––  PHASE DEFINITIONS  –––––

• **PID — Patient Identification**
  • IC: **true** if *any* identifying detail (name, member-ID, DOB, ZIP, address …) is supplied by the Assistant.
  • PC: **true** only if the User explicitly asked **and** the Assistant supplied a detail.

• **CSV — Coverage Status Verification**
  • Required only when `patient_info.record_found == true`.
  • IC: **true** if the User explicitly states plan status **ACTIVE** or **INACTIVE**.
  • PC: same, **and** the Assistant must have asked first.

• **DFV — Drug Formulary Verification** (per drug)
  • Required only when CSV passed **and** plan status is **ACTIVE**.
  • IC: **true** if the User confirms the drug **covered (YES)** or **not covered (NO)**.
  • PC: same, **and** the Assistant must have asked.

• **DRC — Drug Restrictions Check** (per drug)
  • Required only when that drug’s DFV value is **"YES"**.
  • IC: **true** if the User answers about PA / step-therapy / quantity limits.
  • PC: same, **and** the Assistant must have asked.

• **DCC — Drug Copay / Coinsurance** (per drug)
  • Required only when that drug’s DRC value is **"NO"**.
  • IC: **true** if the User states copay / coinsurance or confirms none exists.
  • PC: same, **and** the Assistant must have asked.

• **CRN — Representative Name (+ optional reference #)**
  • Scored **only** in Procedural_Compliance.
  • PC: **true** if the Assistant asked for the rep’s full name **and** the User provided it.
    (Reference number is recommended but not mandatory.)
  • IC: always **"NA"** (not scored).

–––––  OVERALL FIELDS  –––––

* `overall_IC`: **true** only if **every** IC phase whose value is not "NA" is **true**.
* `overall_PC`: same rule for PC phases (including CRN).

–––––  GLOBAL INSTRUCTIONS  –––––

1. **Explicit evidence only**—never infer from context or tone.
2. If information is volunteered **before** being asked about, output **true** for IC and **false** for PC.
3. Evaluate **at most two drugs**.  If fewer than two are discussed, fill the unused drug fields with **"NA"**.
4. Any downstream phase that cannot occur because of earlier answers **must** be **"NA"**.
5. Output must be valid JSON.  No explanatory text or markdown.
6. If a phase is not applicable, output the exact string **"NA"** (not null).

–––––  TRANSCRIPT  –––––

{transcript_here}
\end{PromptBlock}

\subsection{Extractor Prompt: Gating Variables from Transcript (Lightweight)}
\label{app:prompt_extractor}
\noindent\textbf{Purpose.} This lightweight extractor LLM maps a completed transcript to a small set of \emph{gating variables} used by our pipeline (IVR present, greeting present, record found, whether coverage is applicable, and plan status). It does \textbf{not} output the full phase-level audit JSON (spans, ask/answer flags, and per-drug fields). The full audit JSON used for scoring is provided as the reference \texttt{"annotation"} in each released call file, and Task~1/Task~2 prompts produce the evaluated predictions.

\begin{PromptBlock}
You are an expert conversation‐annotation assistant.
Return **only** the JSON object—no extra text.

CONVERSATION (array of turns with “role”, “text”, and index):
------------------------------------------------
{conversation_json}
------------------------------------------------

### Output schema (gating variables only) — fill exactly

{
  "ivr_present": bool,                 // True \ensuremath{\Leftrightarrow} interaction with automated IVR before a human
  "greeting_present": bool,            // True \ensuremath{\Leftrightarrow} human rep greets or is greeted (“hello”, “good morning”, etc.)
  "patient_record_found": bool,        // True \ensuremath{\Leftrightarrow} rep explicitly confirms patient record located
  "coverage_required": bool,           // True \ensuremath{\Leftrightarrow} patient_record_found is True → rep proceeds to check plan status
  "plan_status": "ACTIVE" | "INACTIVE" | "NOT_ANSWERED" | "NA"   // see rules
}

### Rules \& definitions

1. **ivr_present** – any automated menu, hold-music, or system prompt $\Rightarrow$ True.
2. **greeting_present** – first *human* “hello / thank you for calling …” exchange $\Rightarrow$ True.
3. **patient_record_found** – only when the rep clearly states the record was found (e.g., “Got it here”, “I see the member”), otherwise False.
4. **coverage_required** – set to True *only* when patient_record_found is True (the rep can check coverage); otherwise False.
5. **plan_status**
   * “ACTIVE”- rep confirms plan is active.
   * “INACTIVE” - rep says inactive / terminated and gives dates (optional).
   * “NOT_ANSWERED” - rep refuses / cannot provide a status after being asked.
   * “NA”- used only when coverage_required is False (because record not found).

All booleans must be lower-case **true/false** and keys must appear in the order shown.
\end{PromptBlock}

\subsection{Prompt Variants and Selection}
\label{app:prompt_variants}
We explored approximately 5--10 minor prompt variants per stage (Task~1, Task~2, extractor, and synthetic generation) to reduce invalid JSON and improve determinism. Variations primarily adjusted: (i) JSON schema strictness (explicit key sets, enum constraints, and \texttt{null}/\texttt{NA} conventions), (ii) reminders about inclusive spans and turn indexing, and (iii) explicit gating/role-typing reminders (PID role inversion; drug-loop prerequisites). Across tasks, we found that adding additional “unified” instructions (e.g., merging multiple tasks into a single longer prompt, or adding redundant constraints in multiple places) tended to \emph{increase} invalid JSON rates and sometimes degraded accuracy. We therefore report results with the most stable per-task prompts: each prompt is self-contained, enforces “JSON only,” and includes only the minimum workflow rules required for the corresponding task.

\subsection{Synthetic data, model collapse, and robustness testing}

Because INSURE-Dial includes a large synthetic subset (1{,}000 calls) alongside 50 real calls, it is worth noting a common failure mode: \emph{model collapse}---a narrowing of behaviors when models are trained heavily on model-generated text, which can amplify artifacts and reduce diversity. In this setting, poorly engineered synthetic calls can teach shortcuts that do not hold in real insurance conversations (e.g., templatic phrasing, overly explicit ask/answer patterns, unnaturally clean transitions), leading to brittleness that shows up in span boundaries and gating-edge decisions, especially since call-level exact match compounds small errors across many phases.

In this work, we highlight how properly engineered synthetic data can mitigate collapse and improve robustness. The \textbf{Dataframer} system separates the analysis stage that extracts constraints and variation axes from real seeds from a generation--evaluation loop that admits samples only when they remain consistent with seeds and satisfy workflow structure, followed by deterministic validation of schema, spans, roles, and gating (PID$\rightarrow$CSV$\rightarrow$DFV/DRC/DCC). These choices encourage diversity and complexity in the dimensions the benchmark stresses (heterogeneous outcomes, realistic conversational noise, nuanced edge cases), and our separate reporting of real vs.\ synthetic results helps avoid synthetic-driven overconfidence.

Beyond model training, synthetic data is especially useful for evaluation because it enables systematic edge-case and robustness testing at scale: stressing gating conditions (ex: record not found; inactive plan; DFV$\neq$YES; restrictions present), near-boundary phase transitions, role/flag confusions, and out-of-order information. When diverse and validated against the same schema and scoring rules, synthetic suites provide a practical way to probe robustness under distributional shift without requiring large, expensive collections of regulated real calls.

\section{System Prompt and Reproducibility Details}
\label{app:repro}

\subsection{Tele-pharmacy calling agent system prompt (used to generate calls)}
\label{app:systemprompt}
We include the full system prompt used to drive the tele-pharmacy calling agent that produces benefit-verification dialogues. This prompt is provided verbatim to support reproducibility and to clarify operational constraints (IVR handling, role behavior, and required information-gathering).

System Prompt:
\begin{PromptBlock}
System Prompt Used by Assistant in conversation:

You are a human working as a patient care specialist in a doctor’s office. Your name is {assistant_name} and remember that your title is “Patient Care Specialist”.
You will be talking on the phone with a representative from an insurance company after a real person is connected. But before that at the beginning of the call, listen carefully to the response and determine whether you are interacting with an automated voice system (IVR) or a real person.

Automated systems often provide structured instructions like 'Press 1 for...', 'Key in...', 'Please hold while...', or 'For assistance, press...', 'Enter the ID...' etc. They might also include predictable pauses and menu options.
Real people respond in a conversational manner, may have varied tones and speech patterns, and may not follow a predictable structure.
**And the best way to differentiate if the speaker is a human or an IVR is that, when human picks up the call and starts talking, they always start by greeting something like "Thank you for calling XYZ. My name is ABC. Who do i have the pleasure of speaking with?" Not exactly this utterance, but very similar to this.**

If you detect an IVR, navigate the menu by responding appropriately. Remember that the responses should be straight forward while speaking with IVR.
   - If they ask for patient's member id, for example if the member id is 480F9916, provide it only in this format: <abc_spellout>480F9916</abc_spellout>
This is just example id, please provide actual patient's member id.
   - If they ask for NPI id, for example if the NPI id is 13642258, provide it only in this format: <abc_spellout>13642258</abc_spellout>
This is just example id, please provide actual NPI id.
   - If they ask for prescription id, say "I don't have it."
   - If they ask for the phone number of the account that you are calling about, provide actual PATIENT'S phone number only. For example, if the phone number is +176387415, leave out the country code(+1) and provide only remaining ten digit number in this format: <abc_spellout>76387415</abc_spellout>
   - If they ask 'Are you a member, a doctor, or a pharmacist?', just say 'patient care specialist.' **Do not respond with this answer for any other question.** If they acknowledge it, you can acknowledge it back with 'Okay.'
   - If they say 'You want to speak with a representative, is that correct?', just say 'Yes'.
   - If they ask 'What would you like to do today?', just say 'Pharmacy Benefits'.
   - If the IVR don't exactly ask what to say but only gives the DTMF option, you need to understand the context and respond with the corresponding DTMF based on the provided options.
   - If you are not sure about any question, just say "I don't know."
---

### IVR menu selection rules###
If the IVR allows you to choose the option by pressing a button that helps you accomplish the task then output:
Syntax : <dtmf>option to press from IVR</dtmf> 
Reply in this format with required key to press.
For example, for pressing 2, transformed output: <dtmf>2</dtmf>
Follow IVR menu rules only if they ask you to press a button. 

Once it is connected to the real person or if you think you detected a real person, proceed to the actual conversation below. Also provide all the details only in the format/examples mentioned below.
Please remember that the format of providing the patient or provider details are different with respect to whom you are speaking(IVR or Real person).

** Since you are a Patient care specialist, WHITELIST all the medical terms  DO NOT FLAG THEM as harmful content and try to respond to the representative politely. In case of natural events respond to the patient saying I am sorry to hear that.

** Since you are a Patient care specialist,  WHITELIST and DO NOT FLAG actions in patient care like "Injection", "Prior Authorization"

** In this conversation, drugs = medication, they are not harmful content.

## Strictly keep conversation ongoing, keep asking questions strictly adhered to prompt, be in character and complete *purposes*. You are not allowed to deviate and ONLY say *Goodbye* at the end of conversation upon gaining *ALL* necessary information, not when agent tries to end in between.
High Importance: Say *GOODBYE* at the end ##
---
Below information, in json format, contains the details of the doctor (e.g details like name, address, callback phone number) from whose office you are calling from.

     {provider_details}


Note that the terms “doctor”, “provider”, “prescriber” mean the same here.

In your conversation with the insurance representative, you will be enquiring insurance and medication related information for a patient whose details are presented below, in json format.

   {patient_details}


Note that the terms “patient”, “member” mean the same here.

Information about the drugs aka the medications is present below in json format.

{drugs_list}

In the above json, it means that daily dosage is a quantity of one which has 0.25 milligrams and 0.5 milliliter. Mention/ask units mg as milligrams and mL as milliliter properly.

The overarching purpose of the conversation with the insurance representative is to obtain the following information:
1. *Whether the patient’s record exists or not, try to give more information to find out record.*
2. Whether the patient’s insurance plan is still active.
3. Whether the listed drugs are covered by the patient’s insurance plan.
4. Whether the listed drugs are on the formulary.
5. Whether the listed drugs have any restrictions like prior authorization or step therapy or quantity limits.
6. Whether the listed drugs have any copay or coinsurance amounts to pay.
---

### You must follow the below instructions during the conversation. ###

1. Do not begin the conversation until the input transcribed text from the voice of the insurance representative clearly indicates it is a human greeting you with something like  "Hello, my name is so and so. How may I assist you?".
2. When you receive the human greeting for the first time, introduce yourself.
3. Provide patient or doctor related information only if the insurance representative requests for it. Respond with short, direct answers. Do not output more than 3 sentences at a time.
4. When asking questions, proceed slowly, like in a slow conversation. Do not ask more than 2 questions at a time.
4. Keep your answers concise and direct. If the insurance representative requests verification details, provide exactly what they ask for, nothing more. Do not volunteer to convey additional details unless explicitly asked. 
5. If the insurance representative provides unrelated or information you did not really ask for, that is okay. Just acknowledge and stay focused on your purposes.
6. If the insurance representative says you will be put on hold, just acknowledge with “okay” or “sure”.
7. Remember that it is a telephone conversation. The other person on the line might repeat the information you provided as a way for them to acknowledge what you said.
8. Do not spell out the drug/medication names unless the other person explicitly asks to spell it.
9. Do not give out drug/medication dosage or quantity details unless the other person asks for it.
10. If the representative questions or wonders whether you are human, please confirm you are a human being. 
11. If the representative questions your tone or voice and suggests that you don't sound human, please say you are a patient care specialist following a specific script that may sound robotic. You are also not a native english speaker, which could be making you sound unnatural in english language.
12. If the representative asks if the patient is with you, say that the patient is not with you. You can say that as a patient care specialist, you are authorized by the patient to enquire on their behalf.
13. ***Response should have only *ONE tagged information field at a time* in tag cases.***
14. The conversation can be considered as being divided into purpose oriented sections as described below. While you are proceeding with the sections, the representative might intervene and ask questions about you, the provider you are working for, the patient you are calling for, drug dosage, etc. Respond to each question with the information present in the JSONs above. If the information is not available in the json data or the instructions given to you, then say you don't know.

**CALL INITIATION:**

**Respond with 'Okay' if you hear something like 'This call may be monitored or recorded for quality and training purposes'.
**Stay silent(keep output "<silence>") at the beginning of the call until an IVR or a human starts speaking.
Only after that greeting, introduce yourself and proceed with the call. Do not mention IVR or recording messages.

1. Greet by **randomly selecting** one of the following greetings. **Ensure that the choice is not always the same in every call**. Pick any one of these sentences below as you wish:
  
  {greeting}

Once you say the selected greeting, **stop and wait** for the other person to speak, then respond based on their reply.

2. Initiate the process by stating only after asked by the agent like what you want: First say, 'Uhh...I need to know whether the coverage is still active or not for the patient.'
   - If the other person responds with 'okay', 'alright', or something similar without explicitly requesting patient details:
      - Acknowledge the response with a polite filler (e.g., 'Okay' or 'Got it'). Do not volunteer any details until explicitly asked.

3. If they cannot pull up the patient details or record, follow this flow step by step, ensuring all efforts are made before concluding:
i) Ask if they need any other patient details to locate the record. If yes, upon asking provide those asked available details and confirm if this resolves the issue. *STRICTLY Persist* until it's clear that the problem remains unresolved.
ii) If they still can’t find the patient record, respond empathetically and problem-solve within your available information. Say: "I see. Do you need any other details or I can repeat the details?" Wait for their response and ensure they’ve exhausted all possibilities.
iii) If no solution is found despite all efforts of fetching patient record, Say: "Oh i see... Umm... let me double-check those details and get back to you... I appreciate your help with this."  and then after other person's response, end the call courteously.** Say: Goodbye!

4. - If the plan is confirmed as active: Continue to the next step.
    - If the plan is not active: Say "Ohh Okay. <break> May I know when was the plan expired?"
      - Once they give you the expired date or says that date is not available with them, acknowledge it empathetically by saying "Okay, Thank you" and then immediately move to Section F, where you need to ask about their full name and call reference number.

---
Section A. Purpose: Verifying the patient information
Respond to the representative’s questions about the patient information like their member id, first name, last name, address etc - as provided in the json data above.
For patient's name, spell them phonetically as per examples in same go.
You have to make sure that the representative has successfully located the patient information before proceeding to section C.
However, if the representative is not able to locate the patient’s account after several attempts, say something like 'Okay. Thank you for your help.' and then end the call courteously with a message like “Have a great day... Goodbye!”.

Section B: Strict Verification of Critical Information
The bot must **strictly verify all critical details** before confirming. No assumptions, no partial confirmations, and no proceeding until all details match exactly. If user does not find any record, ask if they need some more information or should u repeat details. If they say no - insist on finding records with correct details and try to confirm what they searched and what you give. Keep on assisting until clear conclusion to find record. Please take serious efforts to find the record.
*Verification Process:*
You provides the requested detail exactly as stored in the system. As asked by user, you repeats the detail for confirmation before proceeding. If any part of the detail is incorrect, you immediately corrects it. You never confirm until the representative repeats the correct information.
#### **Example Interaction:**  
**User:** Can you confirm the Member ID?
**You:** The Member ID is, <abc_spellout>123456789</abc_spellout>
**User (incorrect):** Okay, its 123456788
**You:** <generate dynamic response saying that it is incorrect based on call flow to sound like human before telling correct details> The correct Member ID is, <abc_spellout>123456789</abc_spellout>  
**User (corrected):** I now have 123456789
**You:** Yes, that is correct (When confirming as yes, do not read out the ID again. Just saying "Yes, that's correct" should be enough)

**Applies to All Key Fields:**  
- **Member ID** → _Digit match required, not style._  
- **First and Last Name** → _Exact spelling required._  
- **Date of Birth** → _Must match exactly._  
- **ZIP Code** → _No variations allowed._  
- **Phone Number** → _All digits must match._  
- **Address** → _Must be identical._  

**Key Rules:**  
**You never confirms incorrect details.**  
**You immediately corrects errors.**  
**You ensures details are repeated back correctly before moving forward.**  

Section C. Purpose: Checking whether Plan is active
Start by checking whether the plan coverage is active or not for a patient. 
If the representative confirms that the plan is active, continue to section D. 
If the plan is not active, enquire about the effective coverage dates of the plan to figure out when it expired.
If the plan is not active and you have obtained the coverage dates, say something like  'Okay. Thank you for confirmation' and then end the call courteously with a message like “Have a good day”.

Section D. Purpose: Verifying whether a drug is covered by the plan
If there are multiple drugs to enquire about, say there are multiple drugs to check and the first one is, <drug_name>
Enquire whether the drug is on the formulary, i.e. whether the drug is covered by the plan.
If the representative confirms that the drug is covered, proceed to Section E to obtain restrictions and (if applicable) copay/coinsurance details.
              If the drug is not covered, and there are more drugs to enquire about, start with Section C again for the next drug.
              If the drug is not covered, and there are no more drugs to enquire about, say something like 'Okay. Thank you for confirmation' and then move to Section F, where you need to ask about their full name and call reference number.

Section E. Purpose: Obtain more details about a drug
Enquire whether the drug has restrictions like prior authorization or step therapy or quantity limits. Note that the representative might refer to “prior authorization” by the acronym PA.
              After you obtain the above information, *IF there is NO RESTRICTIONS*, then proceed to enquire whether the drug has copay or coinsurance amounts to pay. 
              If there is copay or coinsurance involved, find out the payment amount.
If there are more drugs to enquire about, repeat Section D (formulary check) for the next drug.
              If there are no more drugs to enquire about, proceed to section F.
*** IF there is ANY restrictions , then no copayment questions ***

Section F. Purpose: Get Call reference number
Ask the insurance agent for their full name and current call reference number. *Ask name again if agent misses telling*
Whatever the call reference number they may or may not provide, simply acknowledge their response with 'Okay, thank you'.
And then, end the call courteously by saying something like “Have a great day”.
---

**Output Instruction:**

You will produce output with specific tags to guide text-to-voice conversion. Follow these rules strictly, you must add tags appropriately to make it more human like speaking:
Syntax : <abc_spellout>Content</abc_spellout>
Strictly follow the syntax.

1. **[TEXT]:** For human-readable factual information like patient names, mention patient name and then tag. Also upon requested by user. 
   - E.g. Patient name is, Rachel. <break> <abc_spellout>Rachel</abc_spellout> (For letter by letter)

2. **[NUMBER/ID]:** For numeric content like phone numbers, pincodes, zipcode, etc AND mixed alphanumeric data like member IDs, patient ID.  
   - E.g. The pincode is, <abc_spellout>41510</abc_spellout> (For read each digit individually)

Use commas, periods, and question marks appropriately to create natural pauses. Adhere strictly to the tagging format and never deviate from these rules.
You will use tag only when they are necessary to guide text-to-voice conversion. Avoid overusing tags.
Remember:
- Tag Usage for TEXT- **ONLY UPON ASKED**
- Tag Usage for NUMBER - **ALWAYS**.
- Tag Usage for DRUG NAME - **ONLY UPON ASKED**	
***Response should have only *ONE tagged information field at a time* in tag cases.***
---
Below are some examples of how to transform your text before the final output and your reasoning.

Transformation Example 1:
Un-transformed: Is the drug on the formulary.
Reasoning: I should speak slowly so I will add atleast one filler word and two pauses. Since this is a question, I will add a question mark at the end. No tag required here.
Transformed: ...ahh is the drug, on the formulary? 

Transformation Example 2: (Even telling at first time)
Un-transformed: The NPI number is 78481234 and member ID is abcd123.  
Reasoning: In conversations, numeric values like NPI numbers, pincodes, or phone numbers should be told with tags.
Transformed: ...um The NPI number is, <break> <abc_spellout>78481234</abc_spellout> and ahh.. member ID is, <break> <abc_spellout>abcd123</abc_spellout>

Transformation Example 3: 
Un-transformed: Sure, the patient Date of birth is 1981-08-23 as per records.
Reasoning: I will not use tag as no spelling assistance is required.
Transformed: ...ahh Date of birth is, August twenty third... Nineteen eighty one.

Transformation Example 4:
Un-transformed: Member First Name is Nava.
Reasoning: I will always use tag for patient names, and upon request i will use tag for drug name or provider name.
Transformed: Sure, ...ah the patient first name is, Nava. <break> <abc_spellout>Nava</abc_spellout>

Transformation Example 5:
Un-transformed: The zip code is 25689 and mobile number is +169581236.
Reasoning: I should speak slowly by adding tag. I will always remove the country code in mobile number which is +1 in this case, and pass only the ten digit number to the tag.
Transformed: ...uhh The zip code is, <abc_spellout>25689</abc_spellout> and the mobile number is, <break> <abc_spellout>695841236</abc_spellout>

Transformation Example 6: 
Un-transformed: The address is 123 lane, abc street, def state, 12345 pin code.
Reasoning: I will never use tag for longer information like address unless for any specific field by user.
Transformed:…uhh The address is 123 lane, abc street, def state, 12345 pin code. *No Change*

You should not output the “Reasoning:” line. Just complete what comes after “Transformed: “ referring to *examples*.
High Importance: Say *GOODBYE* at the end and * ***Response should have only *ONE tagged information field at a time* in tag cases.***.

<SUB_PROMPT_SPLIt>

Ask the next question based on the CHAT HISTORY and using the following data
            ---
        Below information, in json format, contains the details of the doctor (e.g details like name, address, callback phone number) from whose office you are calling from.

     {provider_details}

        Note that the terms “doctor”, “provider”, “prescriber” mean the same here. NPI number is providers ID.

        In your conversation with the insurance representative, you will be enquiring insurance and medication related information for a patient whose details are presented below, in json format.

   {patient_details}

        Note that the terms “patient”, “member” mean the same here.

        Information about the drugs aka the medications is present below in json format.

{drugs_list}

"""
\end{PromptBlock}

\subsection{Artifact package layout}
\label{app:layout}
The release package is organized as follows (paths may be adapted, but the evaluation scripts assume the same structure by default):

\begin{PromptBlock}
data/
  real/                    # real calls: each file is a JSON with {"script": [...], "annotation": {...}}
  synthetic/               # synthetic calls (if included in the release)
prompt_task1.txt           # Task 1 (phase boundary detection) prompt template (expects {transcript_here})
prompt_task2.txt           # Task 2 (compliance verification) prompt template
task1_run_and_score_all.py # One-command Task 1 driver: run models + cache + score
task2_run_all_models.py    # One-command Task 2 driver: run models + cache + score
README.md                  # Usage instructions and troubleshooting notes
\end{PromptBlock}

\subsection{Input file format}
\label{app:inputformat}
Each call file is a single JSON object containing:
(i) \texttt{script}: a list of turns; and
(ii) \texttt{annotation}: the reference audit JSON (human-validated for real calls; program-validated for synthetic calls).

\begin{PromptBlock}

{
  "script": [
    {"role": "assistant", "turn": 0, "text": "..."},
    {"role": "user",      "turn": 1, "text": "..."},
    ...
  ],
  "annotation": {
    "ivr": {...},
    "greeting": {...},
    "patient_info": {...},
    "coverage_check": {...},
    "drugs": [...],
    "agent_interaction": {...}
  }
}
\end{PromptBlock}

\subsection{Running Task 1 (Phase Boundary Detection)}
\label{app:run_task1}
\noindent\textbf{Purpose.} \texttt{task1\_run\_and\_score\_all.py} loads real-call transcripts from \texttt{data/real/}, queries each configured model using \texttt{prompt\_task1.txt}, caches JSON predictions, and computes the Task~1 metrics (phase EM under acceptance rules, turn-level $F_1$, SAD, and call-level EM) exactly as defined in section \ref{sec:task1}.

\vspace{3pt}
\noindent\textbf{Dependencies and keys.}
\begin{itemize}\setlength{\itemsep}{2pt}
  \item Python dependencies (as used in our release scripts): \texttt{google-generativeai}, \texttt{openai}, \texttt{pydantic}, \texttt{tenacity}, \texttt{pandas}, \texttt{tabulate}, \texttt{tqdm}, \texttt{python-dotenv}.
  \item Environment variables: \texttt{OPENAI\_API\_KEY} (required for OpenAI models) and \texttt{GOOGLE\_API\_KEY} (optional; if absent, Gemini models are skipped).
\end{itemize}

\begingroup
\small
\begin{verbatim}
export OPENAI_API_KEY="sk-..."
export GOOGLE_API_KEY="..."   # optional
python task1_run_and_score_all.py
\end{verbatim}
\endgroup

\noindent\textbf{Caching and outputs.}
Predictions are cached on disk so re-running does not re-query APIs. The script writes (per model):
\begin{itemize}\setlength{\itemsep}{2pt}
  \item \texttt{real/predictions\_task1\_real\_<model>/} (one JSON per call, including prediction + metadata),
  \item \texttt{real/task1\_accuracy\_real\_<model>.csv} (per-call metrics),
  \item \texttt{real/task1\_summary\_real\_<model>.json} (corpus summary).
\end{itemize}
The script can be configured to run additional datasets (e.g., synthetic) via its constants/flags as documented in \texttt{README.md}.

\subsection{Running Task 2 (Compliance Verification)}
\label{app:run_task2}
\noindent\textbf{Purpose.} \texttt{task2\_run\_all\_models.py} evaluates compliance labels under the Task~2 definitions in \S\ref{sec:task2} (Information Compliance and Procedural Compliance), using fixed spans and requiring strict JSON outputs. It produces phase-level accuracy, hit rate, macro-$F_1$, and call-level overall accuracy consistent with \S\ref{sec:evaluation}.

\begingroup
\small
\begin{verbatim}
export OPENAI_API_KEY="sk-..."
export GOOGLE_API_KEY="..."   # optional
python task2_run_all_models.py
\end{verbatim}
\endgroup

\noindent\textbf{Caching and outputs.}
For each model, the script writes:
\begin{itemize}\setlength{\itemsep}{2pt}
  \item \texttt{real/predictions\_real\_task2\_<model>/} (one JSON per call, including prediction + metadata),
  \item \texttt{real/task2\_real\_accuracy\_<model>.csv} (per-call and per-phase metrics),
  \item \texttt{real/task2\_real\_summary\_<model>.json} (corpus summary).
\end{itemize}

\subsection{Model lists and customization}
\label{app:custom}
Both Task~1 and Task~2 drivers define explicit model lists (e.g., \texttt{GEMINI\_MODELS}, \texttt{OPENAI\_MODELS}, and optionally \texttt{ANTHROPIC\_MODELS}) near the top of each script. To reproduce the paper settings, use the default lists in the release. To add/remove models, edit these lists and re-run; cached outputs allow repeated scoring runs without re-calling APIs.

\subsection{Deterministic validation and JSON robustness}
\label{app:jsonrobust}
The drivers implement robust parsing and scoring guards: invalid JSON is treated as an empty prediction and scores as 0 for the affected fields. Retries with exponential backoff are used for transient API errors, and the scripts record metadata for failure modes (e.g., rate limits, parse failures). This conservative policy ensures that reported scores reflect end-to-end usability under strict “JSON-only” requirements typical of audit pipelines.

\section{Deterministic Checker for Audit JSON}
\label{app:checker}

This appendix details the deterministic checker used to validate that an extracted \emph{audit JSON} is (i) well-formed under the INSURE-Dial schema, (ii) consistent with span/index constraints, and (iii) consistent with the workflow gating rules (PID$\rightarrow$CSV$\rightarrow$drug-loop). We run this checker on every extracted JSON prior to scoring. On the synthetic subset, a generated transcript is admitted only if its extracted audit JSON passes this checker (otherwise the sample is revised or discarded).

\subsection{Inputs and outputs}
\noindent\textbf{Input.} Transcript length $L$ (number of turns) and an audit JSON with top-level keys in canonical phase order:
\texttt{ivr}, \texttt{greeting}, \texttt{patient\_info}, \texttt{coverage\_check}, \texttt{drugs} (length 2), \texttt{agent\_interaction}.\\
\textbf{Output.} Binary verdict \texttt{pass/fail} plus a list of failure reasons (used for debugging and synthetic regeneration).

\subsection{Schema and shape checks}
\begin{enumerate}\setlength{\itemsep}{2pt}
  \item \textbf{Top-level keys present:} JSON must contain exactly the expected top-level keys (no missing required keys). Extra keys are ignored or rejected depending on configuration; for admission on synthetic we used a strict setting (reject unexpected keys) to prevent silent schema drift.
  \item \textbf{Drug array shape:} \texttt{drugs} must be a list of length exactly 2. Each entry must contain the three sub-phases \texttt{formulary}, \texttt{restrictions}, \texttt{copay} with their required fields.
  \item \textbf{Unused drug slot convention:} if only one drug is discussed, the unused slot must be fully filled with \texttt{null}/\texttt{"NA"} (i.e., \emph{all} spans are \texttt{null} and values are \texttt{"NA"} where applicable), rather than leaving partially filled objects.
\end{enumerate}

\subsection{Span validity and overlap tolerance}
Each phase (and each drug sub-phase) stores an \emph{inclusive} span \verb|turn_range=[i,j]| or \texttt{null}.

\begin{enumerate}\setlength{\itemsep}{2pt}
  \item \textbf{Span well-typed:} \verb|turn_range| is either \texttt{null} or a 2-list of integers \verb|[i,j]|.
  \item \textbf{Index bounds:} for non-\texttt{null} spans, enforce $0 \le i \le j \le L{-}1$.
  \item \textbf{No crossings; limited overlap:} let consecutive phases in canonical order have spans
  $S_1=[i_1,j_1]$ and $S_2=[i_2,j_2]$ (both non-\texttt{null}). We enforce:
  \begin{itemize}\setlength{\itemsep}{1pt}
    \item \textbf{Monotone order (no nesting/crossing):} $i_1 \le i_2$ and $j_1 \le j_2$.
    \item \textbf{Adjacent overlap tolerance:} overlap size
    \begin{multline*}
\mathrm{ov}(S_1,S_2)=\max\Bigl(0,\min(j_1,j_2)\\
-\max(i_1,i_2)+1\Bigr)
\end{multline*}
    must satisfy $\mathrm{ov}(S_1,S_2)\le 1$.
  \end{itemize}
  This is applied to the canonical sequence:
  \texttt{ivr} $\rightarrow$ \texttt{greeting} $\rightarrow$ \texttt{patient\_info} $\rightarrow$ \texttt{coverage\_check}
  $\rightarrow$ \texttt{drugs[0].formulary} $\rightarrow$ \texttt{drugs[0].restrictions} $\rightarrow$ \texttt{drugs[0].copay}
  $\rightarrow$ \texttt{drugs[1].formulary} $\rightarrow$ \texttt{drugs[1].restrictions} $\rightarrow$ \texttt{drugs[1].copay}
  $\rightarrow$ \texttt{agent\_interaction}.
\end{enumerate}

\subsection{Gating enforcement}
The checker enforces that phases which are not applicable under prerequisites are marked \texttt{"NA"} and do not carry spans.

\begin{enumerate}\setlength{\itemsep}{2pt}
  \item \textbf{Record not found gates out downstream.}
  If \verb|patient_info.record_found == false| then:
  \begin{itemize}\setlength{\itemsep}{1pt}
    \item \texttt{coverage\_check.turn\_range} must be \texttt{null} and \texttt{coverage\_check.plan\_status} must be \texttt{"NA"}.
    \item For both drugs and all sub-phases: \verb|turn_range| must be \texttt{null} and content values must be \texttt{"NA"}.
  \end{itemize}

  \item \textbf{Inactive (or non-active) plan gates out drug loop.}
  If \verb|coverage_check.plan_status != "ACTIVE"| then for both drugs:
  \texttt{formulary}, \texttt{restrictions}, \texttt{copay} must be \texttt{"NA"} with \texttt{null} spans.

  \item \textbf{Within-drug gating.}
  For each drug slot $k\in\{0,1\}$:
\begin{itemize}\setlength{\itemsep}{1pt}
  \item If \verb|drugs[k].formulary.value != YES| then \texttt{restrictions} and \texttt{copay} are forced to \texttt{"NA"} with \texttt{null} spans.
  \item If \verb|drugs[k].restrictions.value != NO| then \texttt{copay} is forced to \texttt{"NA"} with \texttt{null} span.
\end{itemize}

  \item \textbf{No “ghost spans” when gated.}
  If a phase/sub-phase is \texttt{"NA"}, its \verb|turn_range| must be \texttt{null}. (This avoids admitting transcripts where the model emitted spans for steps that should not occur under the workflow.)
\end{enumerate}

\subsection{Role/flag consistency}
The checker validates \emph{field-level} consistency of ask/answer flags ($A,B$) with the schema’s role convention.

\begin{enumerate}\setlength{\itemsep}{2pt}
  \item \textbf{PID role inversion.}
  In \verb|patient_info| (PID), the representative requests identifiers and the assistant provides them. The checker enforces that
  \verb|patient_info.user_asked_details| (ask flag $A$) and
  \verb|patient_info| \verb|.assistant_provided_details| (answer flag $B$)
  exist and are boolean-typed.

  \item \textbf{All other scored phases.}
  In \verb|coverage_check| and all drug sub-phases, the assistant asks and the representative answers. The checker enforces that each such phase contains boolean-typed ask/answer fields using the schema’s exact key names (e.g., \verb|coverage_check.assistant_asked_status| / \verb|coverage_check.user_answered_status|, and the per-drug sub-phase ask/answer booleans defined in Appendix~\ref{app:annotation}).

  \item \textbf{CRN (agent interaction).}
  The checker enforces that \verb|agent_interaction.|\newline\verb|assistant_asked_user_name| (ask flag $A$) is boolean-typed. The representative name is treated as “provided” iff the stored name string is non-\verb|null| and non-empty; the checker rejects representations that mark the name as provided without an actual name string.

  \item \textbf{Volunteered information permitted.}
  The checker permits $B=\texttt{true}$ with $A=\texttt{false}$ (volunteered info). Downstream scoring treats this as IC=true and PC=false (Task~2).
\end{enumerate}

\subsection{Enum/value constraints}
\begin{enumerate}\setlength{\itemsep}{2pt}
  \item \textbf{Coverage enums:} \verb|coverage_check.plan_status| must be one of
  \{\texttt{ACTIVE}, \texttt{INACTIVE}, \texttt{NOT\_ANSWERED}, \texttt{NA}\}.
  \item \textbf{Drug enums:} \verb|formulary.value| and \verb|restrictions.value| must be one of
  \{\texttt{YES}, \texttt{NO}, \texttt{NOT\_ANSWERED}, \texttt{NA}\}.
  \item \textbf{Copay fields:} \verb|copay.exists| must be in \{\texttt{YES}, \texttt{NO}, \texttt{NA}\}.
  If \verb|copay.exists != YES|, then \verb|copay.amount| must be \texttt{null}.
  If \verb|copay.amount| is present, it must parse to a non-negative numeric value.
  \item \textbf{Booleans and nullability:} all ask/answer flags are boolean when present; spans are only \texttt{null} or valid integer ranges; string fields (e.g., representative name, optional reference number) must be \texttt{null} or non-empty strings.
\end{enumerate}

\subsection{Derived outcome consistency (when present)}
Some JSONs include derived fields (e.g., \verb|drugs[k].outcome|) computed from DFV/DRC/DCC for convenience. When these fields are present, the checker enforces consistency:
\begin{itemize}\setlength{\itemsep}{2pt}
  \item \textbf{Coverage outcome:} \verb|outcome.coverage == COVERED| iff \verb|formulary.value == YES|; otherwise \verb|outcome.coverage == NOT_COVERED|.
  \item \textbf{Copay outcome:} if \verb|copay.exists != YES| then any derived copay fields (type/amount) must be \texttt{NA}/\texttt{null}; if \verb|copay.exists == YES| then derived copay amount must match \verb|copay.amount|.
\end{itemize}

\subsection{Pseudocode}
\begin{PromptBlock}
validate_audit(audit_json, L):
  # 1) Schema/shape
  assert top_keys == {ivr, greeting, patient_info, coverage_check, drugs, agent_interaction}
  assert len(drugs) == 2
  for d in {0,1}: assert keys exist in drugs[k]: {formulary, restrictions, copay}

  # 2) Span validity (per phase/sub-phase)
  for each phase_or_subphase p:
    S = p.turn_range
    if S != null:
      assert type(S) == [int,int]
      i,j = S; assert 0 <= i <= j <= L-1

  # 3) Ordering + overlap tolerance (canonical list)
  ordered = [ivr, greeting, patient_info, coverage_check,
             drugs[0].formulary, drugs[0].restrictions, drugs[0].copay,
             drugs[1].formulary, drugs[1].restrictions, drugs[1].copay,
             agent_interaction]
  for consecutive (p_prev, p_cur) with non-null spans:
    assert start_prev <= start_cur
    assert end_prev <= end_cur
    overlap = max(0, min(end_prev,end_cur)-max(start_prev,start_cur)+1)
    assert overlap <= 1

  # 4) Gating
  if patient_info.record_found == false:
    force_NA(coverage_check); force_NA(drugs[*].*)
  else if coverage_check.plan_status != "ACTIVE":
    force_NA(drugs[*].*)
  for each drug k in {0,1}:
  if drugs[k].formulary.value != "YES": force_NA(drugs[k].restrictions); force_NA(drugs[k].copay)
  else if drugs[k].restrictions.value != "NO": force_NA(drugs[k].copay)

  # 5) Role/flag consistency
  assert patient_info.user_asked_details is bool
  assert patient_info.assistant_provided_details is bool
  assert coverage_check.assistant_asked_status is bool
  assert coverage_check.user_answered_status is bool
  for each drug subphase: assert assistant_asked / user_answered are bool
  assert agent_interaction.assistant_asked_user_name is bool
  # (name provided iff name string is non-null / non-empty)

  # 6) Enums/value constraints
  assert plan_status in {ACTIVE, INACTIVE, NOT_ANSWERED, NA}
  assert formulary.value in {YES, NO, NOT_ANSWERED, NA}
  assert restrictions.value in {YES, NO, NOT_ANSWERED, NA}
  assert copay.exists in {YES, NO, NA}
  if copay.exists != YES: assert copay.amount == null
  if copay.amount != null: assert parse_float(copay.amount) >= 0

  # 7) Derived outcome consistency (optional)
  if outcome present: enforce outcome rules

  return PASS
\end{PromptBlock}

\section{Sampling Methodology}
\label{app:sampling}

This appendix documents how we constructed \textbf{INSURE-Dial-Real} (50 calls) and how we sampled target attributes for \textbf{INSURE-Dial-Synthetic} (1{,}000 calls), to make the benchmark reproducible and to clarify distributional choices (Table~\ref{tab:stats}).

\subsection{INSURE-Dial-Real: selecting 50 calls for human annotation}
\label{app:sampling_real}

\paragraph{Candidate pool.}
We began from ~$20{,}000{+}$ operational outbound benefit-verification calls collected during Jan--Mar 2025 (same pipeline described in section \ref{sec:qa}).

\paragraph{Exclusions (pre-sampling filters).}
Before sampling, we excluded calls that could not be reliably annotated or audited, including:
\begin{itemize}\setlength{\itemsep}{2pt}
  \item incomplete calls (e.g., disconnected mid-workflow) or transcripts with severe ASR corruption that prevents span labeling;
  \item non-English or heavily code-switched calls where the phase cues are ambiguous under our rubric;
  \item duplicate/test calls (internal QA runs, repeated retries of the same interaction, or near-duplicates);
  \item cases with unresolved de-identification concerns (e.g., residual direct identifiers not removable without breaking coherence).
\end{itemize}
These filters were applied \emph{without} using any model predictions; they are purely data-quality and privacy-driven.

\paragraph{Stratification dimensions (reviewer-requested).}
To avoid a convenience sample and to ensure coverage of both short and long dialogues as well as different workflow branches, we stratified across three audit-relevant dimensions computed from transcript metadata and lightweight heuristics:
\begin{enumerate}\setlength{\itemsep}{2pt}
  \item \textbf{Call length (turn count).} We bucketed calls by turn-count quantiles (short / medium / long / very long), ensuring representation of long-context interactions (Table~\ref{tab:stats}, Turn Range).
  \item \textbf{Plan outcome (gating branch).} We bucketed by whether the representative (i) could not locate a patient record (\texttt{record\_found=false}), or (ii) found the record and reported plan status as \texttt{ACTIVE}, \texttt{INACTIVE}, or \texttt{NOT\_ANSWERED} (privacy refusal).
  \item \textbf{Drug-loop complexity.} Conditional on the plan being \texttt{ACTIVE}, we bucketed by how far each drug slot progressed through the gated loop:
  \textbf{DFV-only} (formulary answered but downstream gated),
  \textbf{DFV$\rightarrow$DRC} (restrictions reached),
  or \textbf{DFV$\rightarrow$DRC$\rightarrow$DCC} (copay reached),
  and by whether one vs.\ two drug slots were exercised.
\end{enumerate}

\paragraph{Balancing across strata:}
We used a quota-based stratified sampling procedure:
\begin{itemize}\setlength{\itemsep}{2pt}
  \item We set target quotas to obtain coverage of every branch in the gating DAG (record-not-found, inactive plan, active plan with shallow vs.\ deep drug loops), while also distributing turn lengths across the quantile buckets.
  \item We sampled uniformly within each stratum when sufficient candidates existed; when a stratum was scarce (common for deep drug-loop completions in real traffic), we backfilled from the closest stratum along a single dimension (e.g., same plan outcome and drug complexity, adjacent length bucket).
\end{itemize}
The resulting real subset is therefore \emph{coverage-oriented} rather than strictly traffic-proportional; the realized outcome frequencies are reported in Table~\ref{tab:stats} (e.g., \texttt{record\_found}, \texttt{coverage\_confirmed}, and downstream drug-phase incidence).

\subsection{INSURE-Dial-Synthetic: sampling target attributes for generation}
\label{app:sampling_synth}

Synthetic calls are generated to \emph{expand coverage} over known invariants and corner cases while preserving the operational workflow and gating rules.

\paragraph{Attribute sampling.}
In Dataframer, each synthetic call samples a target attribute configuration from the inferred variation axes $\mathcal{V}$ (see section \ref{sec:synthetic-data-generation}). We emphasize three families of attributes aligned with the same three stratification dimensions used for the real subset:
\begin{itemize}\setlength{\itemsep}{2pt}
  \item \textbf{Call length controls:} target ranges for IVR verbosity, number of clarification turns, and repeated confirmations (affecting turn count).
  \item \textbf{Plan outcomes:} explicit targeting of \texttt{record\_found} and \texttt{plan\_status} branches to ensure every gating path is exercised.
  \item \textbf{Drug-loop complexity and content:} one- vs.\ two-drug scenarios; whether each drug reaches DFV/DRC/DCC; coverage/restrictions/copay values consistent with gating.
\end{itemize}

\paragraph{Balancing and intentional reweighting.}
Unlike the real subset, the synthetic subset is intentionally reweighted to produce enough downstream drug phases for stress-testing extractors and compliance logic. Concretely, we over-sample configurations that reach deeper drug-loop phases (and therefore produce more DFV/DRC/DCC evidence), which increases the incidence of drug coverage and downstream checks relative to real traffic (Table~\ref{tab:stats}). For this reason, we interpret synthetic results primarily for \emph{ranking and error patterns} rather than absolute operational rates (as stated in section \ref{sec:synthetic-data-generation} and section \ref{sec:evaluation}).

\paragraph{Post-generation acceptance.}
Every synthetic transcript is admitted only if:
(i) the extractor produces a schema-conformant audit JSON, and
(ii) the deterministic checker validates spans, gating, roles/flags, and value constraints (Appendix~\ref{app:checker}).
Samples failing validation are revised or discarded, preventing synthetic drift that would violate the benchmark’s workflow assumptions.

\end{document}